\documentclass[runningheads]{llncs}

\usepackage[mobile]{eccv}

\usepackage{graphicx}
\usepackage{booktabs}

\usepackage{adjustbox}
\usepackage{array}
\usepackage{siunitx}

\usepackage{pifont}

\pdfpxdimen=\dimexpr 1 in/72\relax  %

\newif\ifarxiv
\arxivtrue

\ifarxiv
\usepackage[pagebackref=false,breaklinks=true,colorlinks,citecolor=eccvblue,bookmarks=true,bookmarksnumbered=true]{hyperref}
\hypersetup{
  pdftitle={MouseSIS: A Frames-and-Events Dataset for Space-Time Instance Segmentation of Mice},
  pdfsubject={Computer Vision, Robotics},
  pdfauthor={Friedhelm Hamann, Hanxiong Li, Paul Mieske, Lars Lewejohann, Guillermo Gallego},
  pdfkeywords={Video Instance Segmentation, Mice, Event Cameras, Asynchronous Sensor, High Dynamic Range, High Temporal Resolution}
}
\usepackage[absolute]{textpos}
\else
\usepackage{hyperref}
\fi

\usepackage{orcidlink}

\usepackage{placeins}
\usepackage{graphicx}
\usepackage{caption}
\usepackage{subcaption}
\usepackage{array}
\usepackage{makecell}
\usepackage{multirow}
\usepackage{adjustbox}
\usepackage{rotating}
\usepackage{amsmath}
\usepackage{amsfonts}
\usepackage{amssymb}

\newcommand{\unum}[2]{\multicolumn{1}{c}{\underline{\tablenum[table-format={#1}]{#2}}}}

\newcommand{\bnum}[1]{\bfseries #1}

\newcolumntype{C}[1]{>{\centering\arraybackslash}p{#1}}

\newcommand{\cmark}{\ding{51}}%
\newcommand{\xmark}{\ding{55}}%

\newcommand{\dname}{MouseSIS}  %
\newcommand{\mname}{ModelMixSort}  %
\newcommand{\mnametwo}{EventSeqFormer}  %
\newcommand{\mnamethree}{EventVoxelSeqFormer}  %

\definecolor{light-gray}{gray}{0.75}
\newcommand\gframe[1]{{\color{light-gray}\frame{#1}}}

\begin{document}

\title{MouseSIS: A Frames-and-Events Dataset for\\ Space-Time Instance Segmentation of Mice} 

\titlerunning{MouseSIS: A Dataset for Space-Time Instance Segmentation of Mice}

\ifarxiv
\definecolor{somegray}{gray}{0.5}
\newcommand{\darkgrayed}[1]{\textcolor{somegray}{#1}}
\begin{textblock}{13}(1.5, -0.1)  %
\begin{center}
\darkgrayed{This paper has been accepted for publication at the
European Conference on Computer Vision (ECCV) Workshops, Milan, Italy, 2024.
\copyright Springer
}
\end{center}
\end{textblock}
\fi

\author{Friedhelm Hamann\inst{1,2,3}\orcidlink{0009-0004-8828-6919} \and
Hanxiong Li\inst{1,2,3} \orcidlink{0009-0003-5457-8588}\and
Paul Mieske\inst{2,4}\orcidlink{0000-0003-1513-1946} \and 
Lars Lewejohann\inst{2,4,5}\orcidlink{0000-0002-0202-4351} \and 
Guillermo Gallego\inst{1,2,3,6}\orcidlink{0000-0002-2672-9241}
}

\authorrunning{F.~Hamann et al.}

\institute{Technical University of Berlin, Berlin, Germany \and
Science of Intelligence Excellence Cluster, Berlin, Germany \and 
Robotics Institute Germany, Berlin, Germany \and
Free University of Berlin, Berlin, Germany \and 
German Federal Institute for Risk Assessment, Berlin, Germany \and 
Einstein Center Digital Future, Berlin, Germany
}

\maketitle

\begin{abstract}
Enabled by large annotated datasets, tracking and segmentation of objects in videos has made remarkable progress in recent years.
Despite these advancements, algorithms still struggle under degraded conditions and during fast movements. 
Event cameras are novel sensors with high temporal resolution and high dynamic range that offer promising advantages to address these challenges.
However, annotated data for developing learning-based mask-level tracking algorithms with events is not available.
To this end, we introduce: 
($i$) a new task termed \emph{space-time instance segmentation}, similar to video instance segmentation, whose goal is to segment instances throughout the entire duration of the sensor input (here, the input are quasi-continuous events and optionally aligned frames); 
and ($ii$) \emph{\dname}, a dataset for the new task, containing aligned grayscale frames and events.
It includes annotated ground-truth labels (pixel-level instance segmentation masks) of a group of up to seven freely moving and interacting mice. 
We also provide two reference methods, which show that leveraging event data can consistently improve tracking performance, especially when used in combination with conventional cameras.
The results highlight the potential of event-aided tracking in difficult scenarios. 
We hope our dataset opens the field of event-based video instance segmentation and enables the development of robust tracking algorithms for challenging conditions.
\url{https://github.com/tub-rip/MouseSIS}
\end{abstract}

\section{Introduction}
\label{sec:intro}
Understanding video scenes has been a long-standing goal in computer vision research, with applications ranging from mobile robotics to self-driving vehicles. 
Recently, researchers in ecology and neuroscience have discovered the potential of tracking tools for automated behavior quantification~\cite{Mathis18nature}.
Advances in tracking technology can unlock the analysis of vast amounts of video data, aiding our understanding of biological systems.

While the tracking task has been extensively studied using color (RGB) cameras, methods based on conventional camera images inherit the disadvantages of these sensors. 
Consequently, tracking results are limited under challenging data recording conditions, such as dim light and fast motion.
Several solutions have been proposed to overcome degraded frames, such as ($i$) using high framerate cameras~\cite{Kiani17iccv}, which inherently limit exposure time, or ($ii$) employing denoising and low-light enhancement methods~\cite{Yi23arxiv}, which attempt to recover limited information in frames.

Event cameras are novel vision sensors that record pixel-level brightness changes instead of full images, which offers several advantages \cite{Lichtsteiner08ssc,Gallego20pami}.
These include high temporal resolution and high dynamic range (HDR), making them well-suited for tracking in challenging conditions \cite{Gehrig19ijcv}.
The use of event-based cameras for tracking has been extensively explored, but primarily for the relatively simple task of single-object bounding-box tracking (see~\cref{tab:related:dataset_survey}). 
Consequently, event-based tracking solutions lag behind their counterparts for conventional cameras, which have developed algorithms capable of far more fine-grained tracking of multiple instances at a mask level.
\begin{figure*}[t]
\centering
\begin{subfigure}[c]{0.48\linewidth}
    \centering
    \includegraphics[trim={0 720px 1280px 0},clip, width=\linewidth]{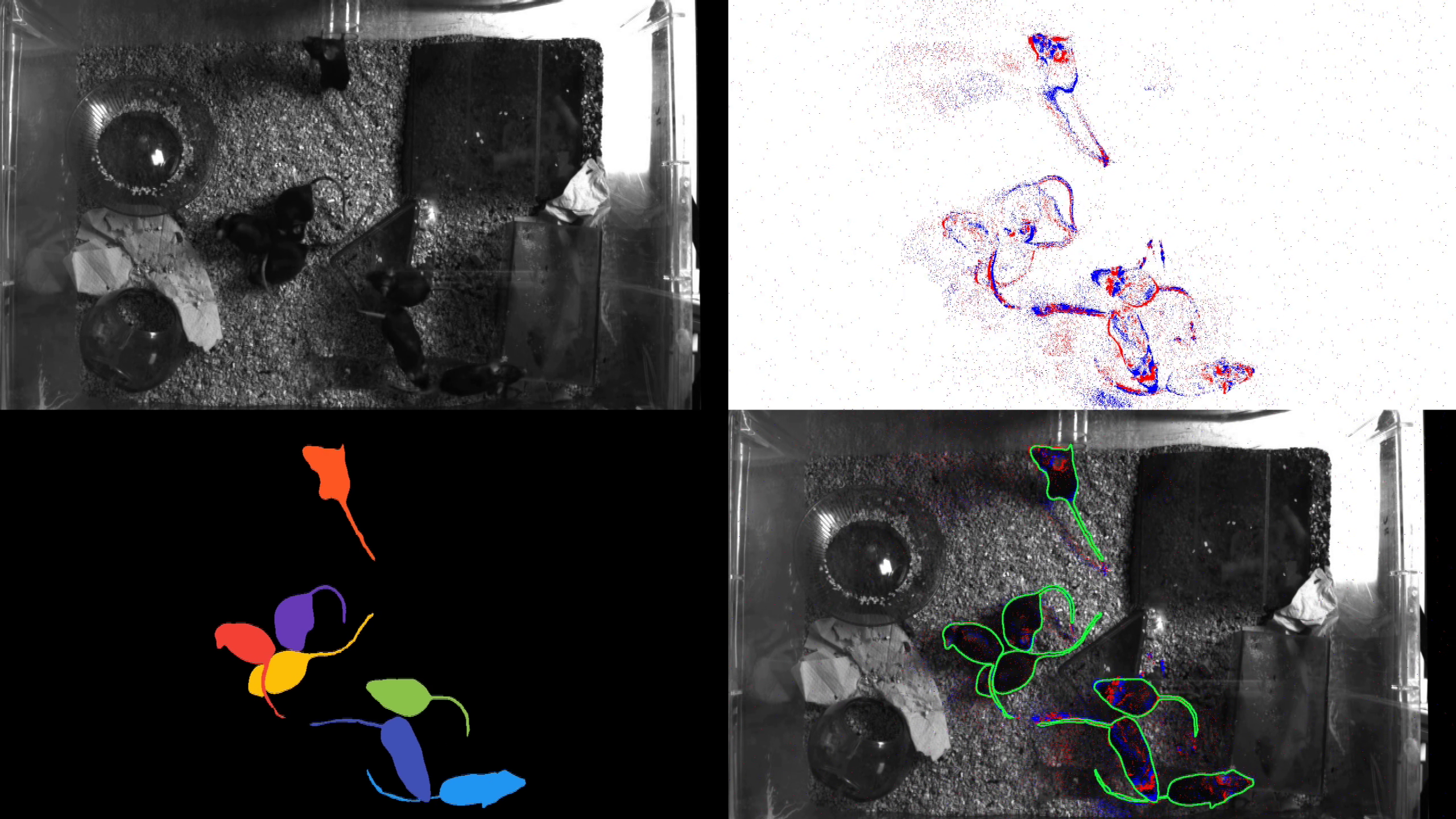}
    \caption{Grayscale frame}
    \label{fig:data:frames}
\end{subfigure}
\,
\begin{subfigure}[c]{0.48\linewidth}
    \centering
    \gframe{\includegraphics[trim={1280px 720px 0 0},clip, width=\linewidth]{images/fig_teaser/placeholder.png}}
    \caption{Events}
    \label{fig:data:events}
\end{subfigure}\\[0.5ex]

\begin{subfigure}[c]{0.48\linewidth}
    \centering
    \includegraphics[trim={0 0 1280px 720px},clip, width=\linewidth]{images/fig_teaser/placeholder.png}
    \caption{Ground truth (GT) segmentation masks}
    \label{fig:data:gt}
\end{subfigure}
\,
\begin{subfigure}[c]{0.48\linewidth}
    \centering
    \includegraphics[trim={1280px 0 0 720px},clip, width=\linewidth]{images/fig_teaser/placeholder.png}
    \caption{Overlaid GT boundaries, events and frames}
    \label{fig:data:overlay}
\end{subfigure}
	\caption{
The dataset contains high-definition (a) frames, (b) events, and (c) instance masks, which are consistent throughout a video.
(d) As the overlay of them shows, events and frames are pixel-level aligned.
Consequently, the masks are valid for both modalities.
}
\label{fig:eyecatcher}
\end{figure*}
Advances in frame-based tracking have largely been unlocked by the availability of more and better-annotated data \cite{Dendorfer20arxiv,Voigtlaender19cvpr}.
The lack of similarly annotated event data is a significant obstacle to the adoption of tasks like video instance segmentation in the event-based domain.

To tackle these central issues, we present a class-guided tracking dataset containing multiple instances with pixel-accurate annotations, similar to the tasks of ``Video instance segmentation''~\cite{Yang19iccv} or ``Multi-object Tracking and Segmentation''~\cite{Voigtlaender19cvpr}.
As events are quasi-time-continuous we term the new task \emph{Space-time instance segmentation (SIS)}.
The targets (moving objects to be tracked in the scene) are up to seven mice in a cage, recorded from a top view (see \cref{fig:eyecatcher,fig:recording_system}).
The dataset contains 33 videos with an average duration of $\approx20$ seconds, which we recorded using a beamsplitter system.
This setup delivers pixel-aligned frame and event data, allowing the exploration of performance trade-offs between the two modalities, as well as their combination.
The sequences contain uneven illumination, and cases of occlusion and crossings among the targets.
We provide high-quality ground-truth (GT) instance masks with consistent identifiers (IDs) for all mice throughout the videos, with a total number of roughly $75000$ instance masks.

In addition, we introduce two reference methods to evaluate our dataset.
The first method builds on a classical tracking-by-detection approach, where we combine several pre-trained models to enhance the base scheme.
Events are converted into reconstructed images via E2VID~\cite{Rebecq19pami}, which are passed to an object detector combined with segment-anything (SAM~\cite{Kirillov23iccv}) to deliver instance masks based on each modality (frames or reconstructed frames).
The per-timestep masks are linked to tracks in an online manner similar to the SORT~\cite{Bewley16icip} algorithm.
The second method builds on SeqFormer~\cite{Wu22eccv}, providing end-to-end learned video instance segmentation capabilities. 
This method is fed with E2VID frames in combination with the grayscale frames.

The results show that our dataset enables effective spatio-temporal instance segmentation using events and frames.
For the tracking-by-detection method, the models using event data consistently outperform the frame-only methods indicating that event data can improve tracking results.
Furthermore, it shows the challenges in our dataset with different contrast thresholds, challenging occlusions, and illumination conditions.
In summary, our contributions are:
\begin{enumerate}
    \item A novel dataset for \emph{space-time instance segmentation} of mice with aligned events and frames.
    We provide $\approx 640s$ of annotated data with 157 spatio-temporal instances yielding a total of 75000 binary masks.
    \item Two reference methods for our new SIS task: A tracking-by-detection-based method integrating several pre-trained components, and an end-to-end learned transformer-based model.
    \item Extensive evaluation of our dataset using the two introduced methods, with results indicating the improved tracking results by including events and showing the challenges of our dataset.
\end{enumerate}

To the best of our knowledge, this is the first publicly available dataset for event-based class-guided mask tracking.
Our dataset and method contribute to a fine-grained scene understanding with event cameras. 
Furthermore, our work contributes to the application of computer vision to a broader scientific domain, such as biology, which is an emergent trend in the field, as exemplified by a recent series of %
workshops\footnote{\url{https://www.cv4animals.com/}}. 
We hope our work opens avenues for robust tracking algorithms in a wider variety of visual conditions.

\section{Related Work}

\begin{table}[t]
\centering
\caption{
Comparison of event-based object tracking datasets.
We only considered real (i.e., non-synthetic) event datasets.
The column ``Aligned'' states whether the frames are pixel-level aligned with the event data or not.
If events and frames have different spatial resolutions, the column ``Resolution'' denotes event camera's value (in px).
}
\label{tab:related:dataset_survey}
\adjustbox{max width=\linewidth}{
\setlength{\tabcolsep}{3pt}
\begin{tabular}{lC{2cm}C{1cm}C{1.5cm}C{3cm}C{3cm}}

\toprule
\textbf{Name}  & \textbf{Resolution} & \textbf{Frames} & \textbf{Aligned} & \textbf{Pixel-level masks} & \textbf{Multiple objects} \\
\midrule

Ulster\cite{Liu16iscas}    & $240 \times 180$  & \cmark  & \cmark & \xmark & \xmark \\
EED~\cite{Mitrokhin18iros} & $240 \times 180$  & \cmark  & \cmark & \xmark & \xmark \\
FE10~\cite{Zhang21iccv}    & $346 \times 260$  & \cmark  & \cmark & \xmark & \xmark \\
VisEvent~\cite{Wang23tcyb}  & $346 \times 260$  & \cmark  & \cmark & \xmark & \xmark \\
COESOT~\cite{Li22wacv}     & $346 \times 260$  & \cmark  & \cmark & \xmark & \xmark \\
EventVOT~\cite{Wang24cvpr} & $1280 \times 720$ & \xmark  & --   & \xmark & \xmark \\
CRSOT~\cite{Zhu24arxiv}  & $1280 \times 800$ & \cmark  & \xmark & \xmark & \xmark \\
DSEC-Detection~\cite{Gehrig24nature} & $640 \times 480 $ & \cmark & \xmark & \xmark & \cmark \\

\midrule

\dname (Ours) & $1280 \times 720$         & \cmark  & \cmark      & \cmark      & \cmark  \\

\bottomrule
\end{tabular}
}
\end{table}

\subsection{Frame-based Object Tracking}
Object tracking has made rapid advances with the availability of deep networks and annotated data, and quickly split up into a variety of task formulations motivated by different end applications.
From the task perspective, a useful taxonomy is dividing tasks into ``exemplar-guided'' and ``class-guided''~\cite{Athar23wacv}.
The first describes tracking methods with an initial cue, e.g. a bounding box in the first frame.
The latter refers to methods that track all instances of a specified set of classes within the video.
We can further divide it into the frame-level representation, usually a bounding box or a pixel-level mask.
In this work, we concentrate on class-guided tracking of pixel-level masks, which is often termed Video Instance Segmentation~\cite{Yang19iccv,Voigtlaender19cvpr}.

In the recent past, we have seen an astonishing increase in the granularity of tracking results.
With the advent of learned object detectors, tracking has often been tackled with a tracking-by-detection approach, where per-frame detections were propagated (e.g., with a Kalman filter plus a linear motion model) and matched based on an association cost~\cite{Bewley16icip,Zhang22eccv_track}.
Advancements were driven by better detectors and bigger, more challenging multi-object tracking datasets~\cite{Leal15arxiv,Milan16arxiv,Dendorfer20arxiv,Yu20cvpr,Sun22cvpr}.
End-to-end learned approaches became feasible but proved to need large amounts of annotated data~\cite{Meinhardt22cvpr}.
This trend has continued but shifted towards more precise mask-annotations~\cite{Xu18arxiv,Athar23wacv}.
A trend not yet picked up by the event community, demanding datasets to close that gap.

One axis of difficulty is the above-mentioned granularity of the representation and number of classes; 
other axes are the visual conditions and motion profiles.
Much less attention has been directed to these areas, which are crucial for many real-world applications.
The work of \cite{Kiani17iccv} proposes to use a high-framerate camera, and \cite{Yu20cvpr} recovers information from frames.
However, all frame-based methods are fundamentally limited by the capabilities of the sensor.

\subsection{Event-based Vision}

An appealing solution to the aforementioned problems in challenging scenarios consists of leveraging the advantages of novel visual sensors called event cameras.
Since the seminal work~\cite{Lichtsteiner08ssc} event cameras have found increasing interest in computer vision and robotics research~\cite{Gallego20pami,Shiba24pami,Hamann24eccv}.
Due to their high dynamic range (HDR) and generally robust imaging under impaired illumination conditions, they have been explored in areas like computational imaging~\cite{Liang24cvpr}, action localization~\cite{Hamann24cvpr} and visual odometry \cite{Rebecq17ral,Guo24tro}.
The results suggest that tracking systems can also directly profit from using this new vision sensor.

Many works are interested in the combination of aligned event data and frames.
A solution is the usage of a camera that directly provides frames and events, like the Dynamic and Active-Pixel Vision Sensor (DAVIS) \cite{Brandli14ssc,Taverni18tcsii}.
However, its spatial resolution is limited, and therefore researchers have customized optical setups like beamsplitter systems~\cite{Hidalgo22cvpr,Hamann22icprvaib,Wang23arxiv_penn,Shiba23pami}, which align the optical axes of a frame and an event camera.
We record our dataset with a beamsplitter system to ($i$) be able to combine both modalities for using the best of both worlds and ($ii$) be able to quantify the trade-off between the modalities.

\subsection{Event-based Tracking Datasets}

Considerably fewer datasets exist for event-based vision compared to those for conventional/standard cameras.
The biggest annotated datasets are available in the automotive context~\cite{Tournemire20arxiv,Perot20neurips,Verma24cvpr} for classification and detection tasks.
{However, they do not provide temporally consistent IDs for objects.}
\Cref{tab:related:dataset_survey} shows an overview of event-based datasets, revealing that most existing datasets are only for single object tracking (exemplar-guided, bounding box).
{Only the DSEC-Detection dataset~\cite{Gehrig24nature}, which is an extension of the DSEC driving dataset, contains multiple object tracking annotations.}
Some of the early datasets provide aligned frames and events from a DAVIS camera but therefore have a low spatial resolution (240$\times$180 px for the DAVIS240C \cite{Brandli14ssc}, or 346$\times$260 px for the DAVIS346 \cite{Taverni18tcsii} ).
Notably, none of the listed datasets provides pixel-level masks.
In contrast, our dataset is of high spatial resolution (1 Megapixel) and contains pixel-level aligned frames and events, and multiple instances annotated with accurate masks.

\section{Dataset}
\label{sec:dataset}
\begin{figure}[t]
    \centering
\begin{subfigure}{0.58\linewidth}
  \centering
  \includegraphics[width=\linewidth]{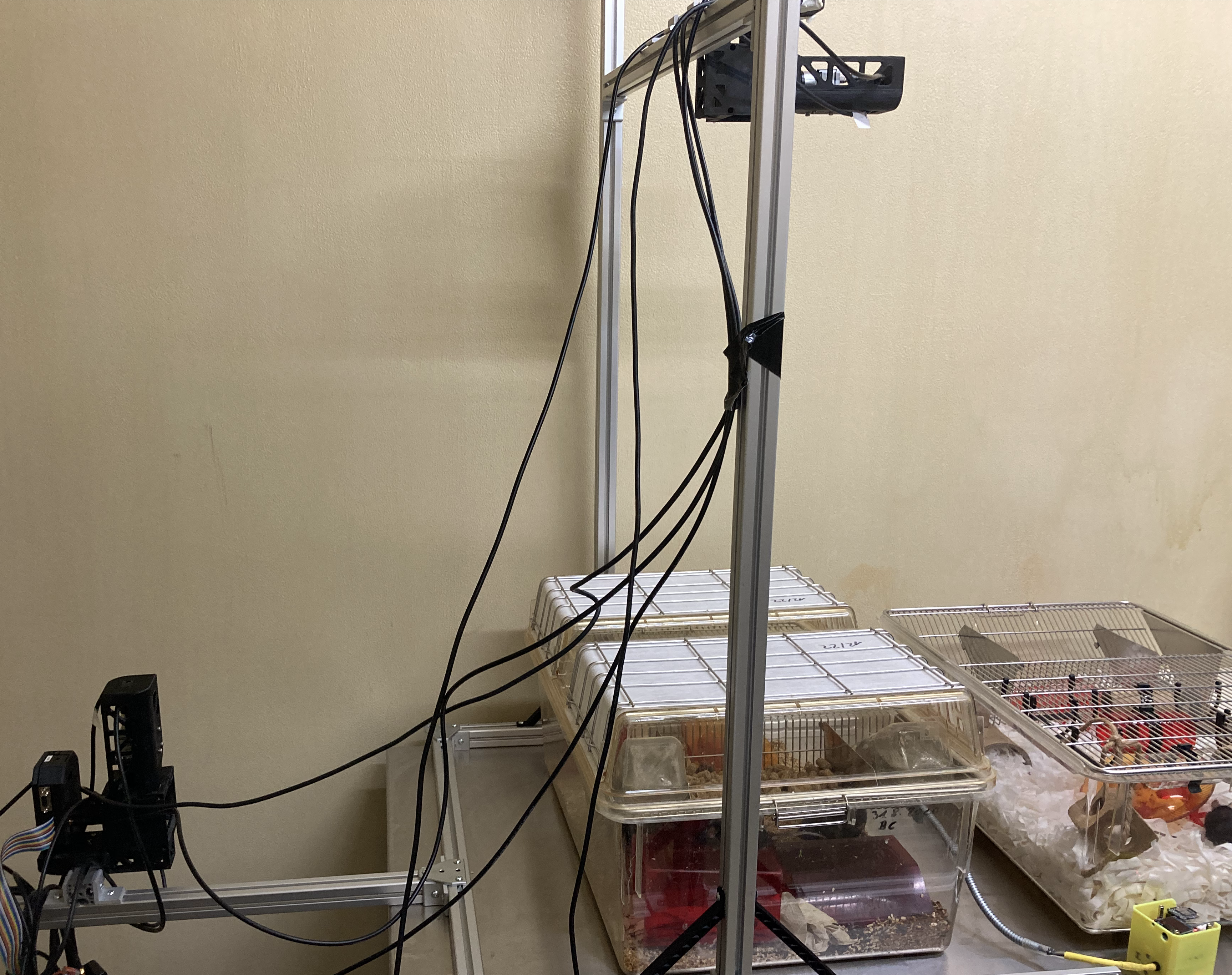}
  \caption{Recording setup: top and front beamsplitters.\label{fig:dataset:heatmap_annotations}}
\end{subfigure} %
\begin{subfigure}{0.38\linewidth}
  \centering
 \includegraphics[trim={9cm 0cm 9cm 0},clip,width=\linewidth]{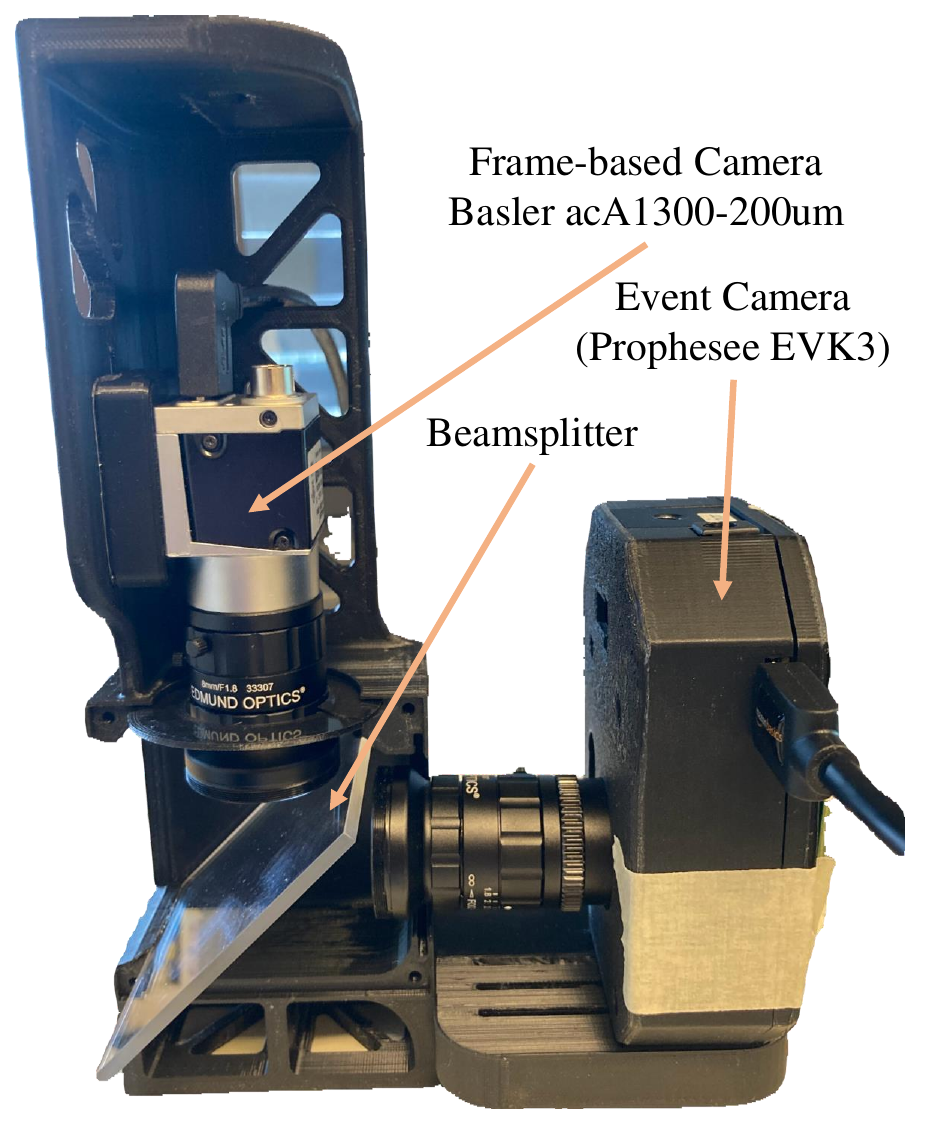}
  \caption{The beamsplitter system.\label{fig:beamsplitter}}
\end{subfigure}
    \caption{(a) Our recording setup consists of several hardware-synchronized cameras.
    Images from the top view were used for the \dname{} dataset.
(b) The beamsplitter system for spatial alignment of frames and events.}
    \label{fig:recording_system}
\end{figure}

\begin{table}[!ht]
  \centering
  \caption{
  Statistics for the \dname{} dataset.
  Values in columns \emph{bias on} and \emph{bias off} are defined by the difference to the middle value of 80 (Prophesee camera).
  \emph{\#Masks} is the number of instance masks, e.g. per mouse in a frame count is increased by one.
  \emph{\#Objects} is the maximum number of mice simultaneously in the field of view.
  }
  \label{tab:data:statistics}
  \adjustbox{max width=\linewidth}{
  \setlength{\tabcolsep}{3pt}  
  \begin{tabular}{l|rrrrrrrr}
    \toprule
   & \textbf{Video ID} & \textbf{Duration} [s] & \textbf{Brightness} [lux] & \textbf{Exposure time} [ms] & \textbf{Bias on} & \textbf{Bias off} & \#\textbf{Masks} & \#\textbf{Objects} \\
  \midrule
  \multirow{22}{*}{ \rotatebox{90}{\makecell{Training set}} } 
  & 2  & 20.03 & 1100 & 16  & -28 &  35 & 2703 & 5 \\
  & 5  & 17.16 & 1100 & 16  & -28 &  35 & 3084 & 6 \\
  & 6  & 20.03 & 1100 & 16  & -28 &  35 & 3092 & 6 \\
  & 8  & 20.03 & 1100 & 16  & -40 &  50 & 4200 & 7 \\
  & 9  & 20.03 & 1100 & 16  & -40 &  50 & 4200 & 7 \\
  & 11 & 20.03 & 1100 & 16  & -40 &  50 & 4200 & 7 \\
  & 13 & 20.03 & 1100 & 16  & -40 &  50 & 1695 & 5 \\
  & 14 & 20.03 & 1100 & 16  & -40 &  50 & 1800 & 3 \\
  & 15 & 20.03 & 1100 & 20  & -40 &  50 & 1640 & 3 \\
  & 17 & 20.03 & 1100 & 20  & -40 &  50 & 600  & 1 \\
  & 18 & 17.19 & 1100 & 20  & -40 &  50 & 380  & 2 \\
  & 19 & 20.03 & 660  & 20  & -40 &  50 & 1800 & 3 \\
  & 20 & 20.03 & 660  & 20  & -40 &  50 & 1679 & 5 \\
  & 21 & 20.03 & 660  & 20  & -40 &  50 & 1200 & 2 \\
  & 23 & 20.03 & 660  & 20  & -40 &  50 & 2846 & 7 \\
  & 24 & 20.03 & 660  & 20  & -40 &  50 & 3063 & 6 \\
  & 27 & 20.03 & 660  & 20  & -40 &  50 & 1546 & 4 \\
  & 29 & 20.03 & 660  & 20  & -40 &  50 & 1532 & 4 \\
  & 30 & 20.03 & 660  & 20  & -40 &  50 & 1285 & 4 \\
  & 31 & 20.03 & 660  & 20  & -40 &  50 & 2744 & 6 \\
  & 33 & 25.41 & 660  & 20  & -40 &  50 & 2287 & 4 \\
  & \textbf{Sum} &420.30 & &   & & &  47576& 97\\
  \midrule
  
  \multirow{5}{*}{ \rotatebox{90}{\makecell{Validation}} } 
  & 3  & 20.06 & 1100 & 16  & -28 & 35 & 2965 & 5 \\
  & 4  & 20.03 & 1100 & 16  & -28 & 35 & 3600 & 6 \\
  & 12 & 20.26 & 1100 & 16  & -40 & 50 & 4162 & 7 \\
  & 25 & 15.08 & 660  & 20  & -40 & 50 & 2496 & 7 \\
  & \textbf{Sum} & 75.43& &   & & & 13223 & 25\\
  \midrule
  
  \multirow{9}{*}{ \rotatebox{90}{\makecell{Test set}} } 
  & 1  & 20.03 & 1100 &  16  & -28 & 35 & 2400 & 4 \\
  & 7  & 14.55 & 1100 &  16  & -28 & 35 & 2586 & 6 \\
  & 10 & 13.54 & 1100 &  16  & -40 & 50 & 2786 & 7 \\
  & 16 & 12.33 & 1100 &  20  & -40 & 50 &  882 & 3 \\
  & 22 & 20.03 &  660 &  20  & -40 & 50 & 1202 & 5 \\
  & 26 & 20.03 &  660 &  20  & -40 & 50 & 1800 & 3 \\
  & 28 & 21.50 &  660 &  20  & -40 & 50 & 1277 & 4 \\
  & 32 & 20.03 &  660 &  20  & -40 & 50 & 1800 & 3 \\
  & \textbf{Sum} &142.04 & &   & & & 14733 & 35\\
  \midrule
  & \textbf{Total} &637.79& &   & & & 75532 & 157\\
  \bottomrule
  \end{tabular}
  }
  \end{table}
\subsection{Dataset Creation}

\subsubsection{Scene setup.}
Our dataset was recorded using a stationary camera system from the top view of a mouse cage on 23.05.2023 with a beamsplitter system combining a frame and an event camera (see~\cref{fig:recording_system}).
All recordings were done in the early morning hours to capture high mouse activity.
The animals were from the mouse strain C57BL/6J.
The setup comprised two Type 4 cage systems connected by a small tunnel, allowing the mice to move freely between the two enclosures.
Several mice were present, able to enter and exit the cages at will.

The camera view included one of the mouse housings shown in \cref{fig:recording_system}, measuring $590 \times 380 \times 200$ mm.
Within the field of view of the cameras, the housing contained several semi-transparent shelters and a spinning wheel located in one corner.
During recordings, the mice interact with these objects, occasionally causing occlusions.
The illumination was intentionally uneven, with a light source placed on one side of the setup.
This uneven illumination created challenging conditions for frame-based cameras, resulting in areas of both overexposure and underexposure within the scene.

\subsubsection{Camera system.}
The setup used for recording was a beamsplitter system, with a modified design based on the open-source system \cite{Hidalgo22cvpr}.
It incorporated a Prophesee EVK3 Gen4.1 event camera (1280 $\times$ 720 px) and a Basler acA1300-200um grayscale camera (1280 $\times$ 1024 px).

Both cameras were equipped with an \emph{EO 8mm UC Series} lens, with a focal length of 8 mm and an aperture set to f/2.8.
Data was recorded under two different settings, varying illumination conditions, exposure times, and event camera bias values, as detailed in \cref{tab:data:statistics}.
Frames were captured at a rate of 30 fps defined by the external trigger signal.
The cameras were hardware-synchronized using a custom trigger box~\cite{Hamann24github}.

Although the beamsplitter provided good alignment of the optical axis, additional calibration was necessary for pixel-level alignment.
We used a checkerboard pattern to extract keypoint matches and determine the homography between the two camera views.
This homography was then used to warp frames into the event coordinates.
Consequently, the final paired data had a resolution of 1280 $\times$ 720 px and was accurately aligned (see \cref{fig:data:overlay}).

\subsubsection{Recorded data.}
We recorded roughly 30 minutes of raw video data with several interruptions.
The data contains two brightness levels (1100 lux, 660 lux) and accordingly adjusted exposure time of the frame-based camera (16 ms, 20 ms).
In a few samples (independent of the illumination) the contrast thresholds are more sensitive, leading to a large amount of noise events.

In addition to the top view, we provide data from the front view seen in \cref{fig:recording_system} (left).
The data is synchronized with the system described before and we provide a 3D calibration performed with kalibr~\cite{Oth13cvpr} and April tags.
The front-view data is not annotated or used in our methods but could be used in future work, e.g., to improve tracking.

\subsection{Dataset Task and Structure}

\subsubsection{Task.}
Our goal is to accurately track moving objects within a data stream, with a specific focus on capturing the precise motion of each mouse throughout a video.
We term this task \emph{space-time instance segmentation} (SIS) similarly to video instance segmentation~\cite{Yang19iccv} but with quasi-continuous events as input.

Given a data stream (e.g., classical video, an event stream, or both), we aim to predict the binary segmentation mask of each object in our ``mouse'' class over time, corresponding to fixed discrete ground truth timestamps.
These timestamps align with the 30 Hz frames in our dataset, mirroring traditional frame-based tasks.

Although the quasi-continuous event stream provides much higher temporal resolution, the prediction task is defined by the frame timestamps.
We maintain the IDs of mice when there is a temporary occlusion through one of the shelters within the scene, however, we assign new IDs to mice exiting and re-entering through the tunnel.

\subsubsection{Structure.}
From the raw recordings, we selected 33 sequences of roughly 20 seconds for annotation.
\Cref{tab:data:statistics} shows an overview of the dataset.
The sequences were annotated in a semi-automated manner, with all manual steps performed by a professional annotation service on the warped grayscale frames.

Initially, we manually annotated frames at 0.5-second intervals, corresponding to every 15 frames.
Then, we used XMem~\cite{Cheng22cvpr}, a video object segmentation model, to propagate each instance mask 8 frames forward and 7 frames backward.
In the final step, we manually corrected the predictions.
We found that fine details of the limbs were often not propagated correctly, and most of the masks predicted by XMem required correction.

The dataset was randomly split into training, validation, and test subsets with an approximate ratio of 70:10:20, respectively, as indicated in \cref{tab:data:statistics}.

\section{Methods}

\subsection{\mname}
\begin{figure*}[t]
	\centering  %
    {\includegraphics[trim={5cm 4cm 5cm 2cm},clip, width=\linewidth]{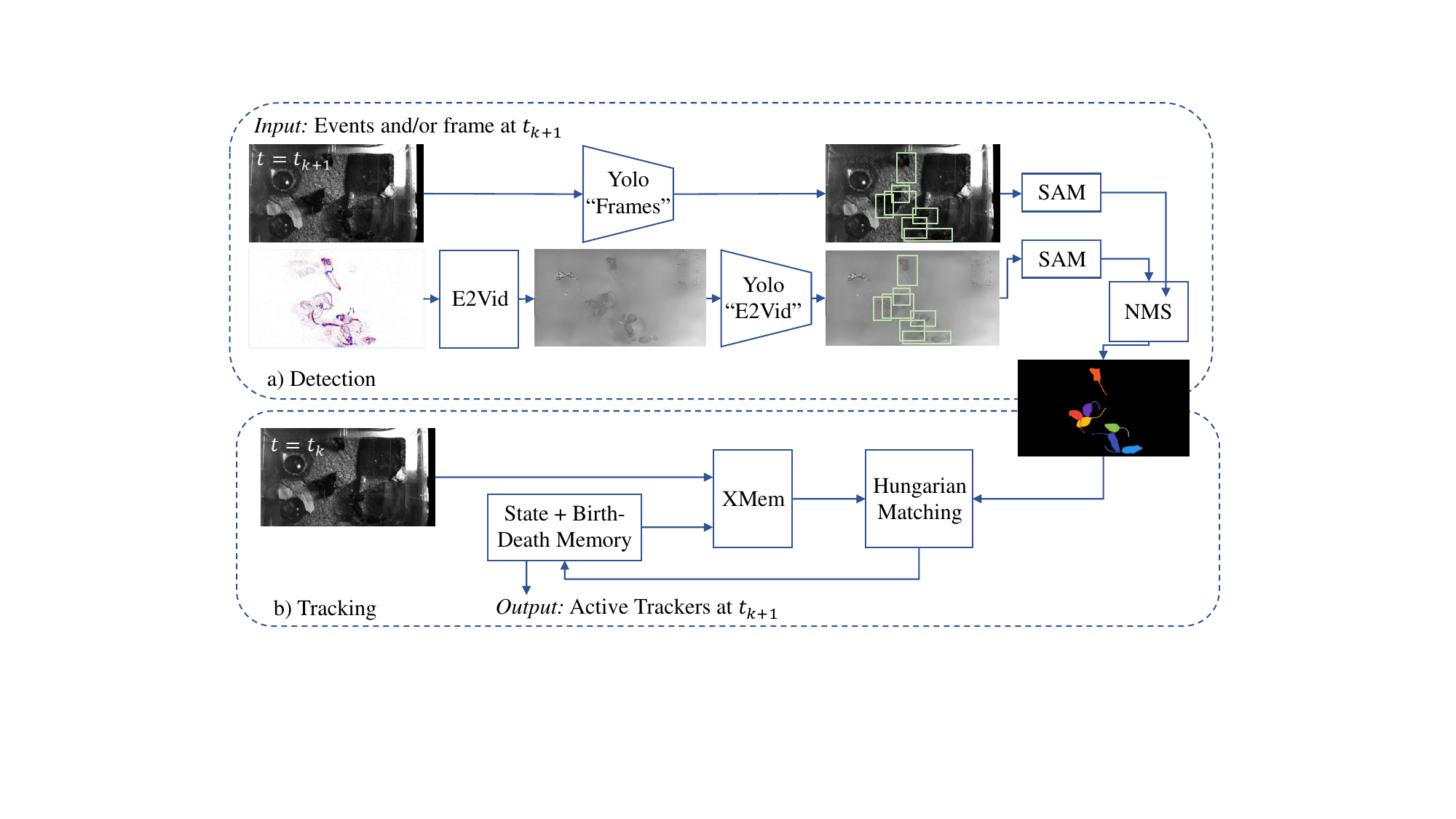}}
	\caption{\emph{Method 1 overview (ModelMixSort).}
 Our method uses a tracking-by-detection approach. (a)
for each frame and the according events, we extract two sets of boxes which are used as box prompts for SAM.
(b) The instance masks are matched to trackers by predicting the current tracker by one timestep and matching it with the detections.
 \label{fig:method:sort}}
\end{figure*}
We provide a reference method, termed \mname, based on a classical tracking-by-detection approach~\cite{Bewley16icip} that incorporates several recent pre-trained models as building blocks.
\Cref{fig:method:sort} provides an overview of the method, which uses the following components:

\begin{itemize}
    \item \emph{E2VID}~\cite{Rebecq19pami}: A model that converts events into grayscale images.
    \item \emph{YOLOv8}~\cite{Jocher23github} (trained on all training set images): An object detection model.
    \item \emph{YOLOv8}~\cite{Jocher23github} (trained on images reconstructed from events at frame timestamps using E2VID): Another object detection model.
    \item \emph{Segment Anything}~\cite{Kirillov23iccv}: A promptable model that segments any object in a scene.
    \item \emph{XMem}~\cite{Cheng22cvpr}: A video object segmentation model that uses a binary mask and a frame as input to predict the object's position in the next frame.
\end{itemize}

The tracking-by-detection approach consists of two steps.
First, the detection module (a)) is applied to each frame and the respective events between the current and previous frames.
Events are converted into image representations by E2VID. 
Both grayscale and E2VID images are processed using their respective object detectors to obtain bounding boxes, which serve as prompts for SAM to generate two sets of detected instance masks.
These sets are then merged using non-maximum suppression (NMS).

In the tracking module (b)), per-frame detections are associated in an online manner. 
This process is similar to~\cite{Bewley16icip}, with the key difference being that the tracker's state is represented by a binary mask, and the prediction step is carried out using XMem instead of Kalman filters.
The matching cost for Hungarian matching is the mask IoU between instances and trackers.
The online tracking algorithm initializes new trackers whenever a detection cannot be matched.
If a tracker had three consecutive detections it is marked as active, if it was not detected more than one time it is discarded.

Both detection models were trained for 50 epochs on an A4000 Nvidia GPU. 

\subsection{\mnametwo}
\begin{figure*}[t]
	\centering  %
    {\includegraphics[trim={4cm 2cm 4cm 1cm},clip, width=\linewidth]{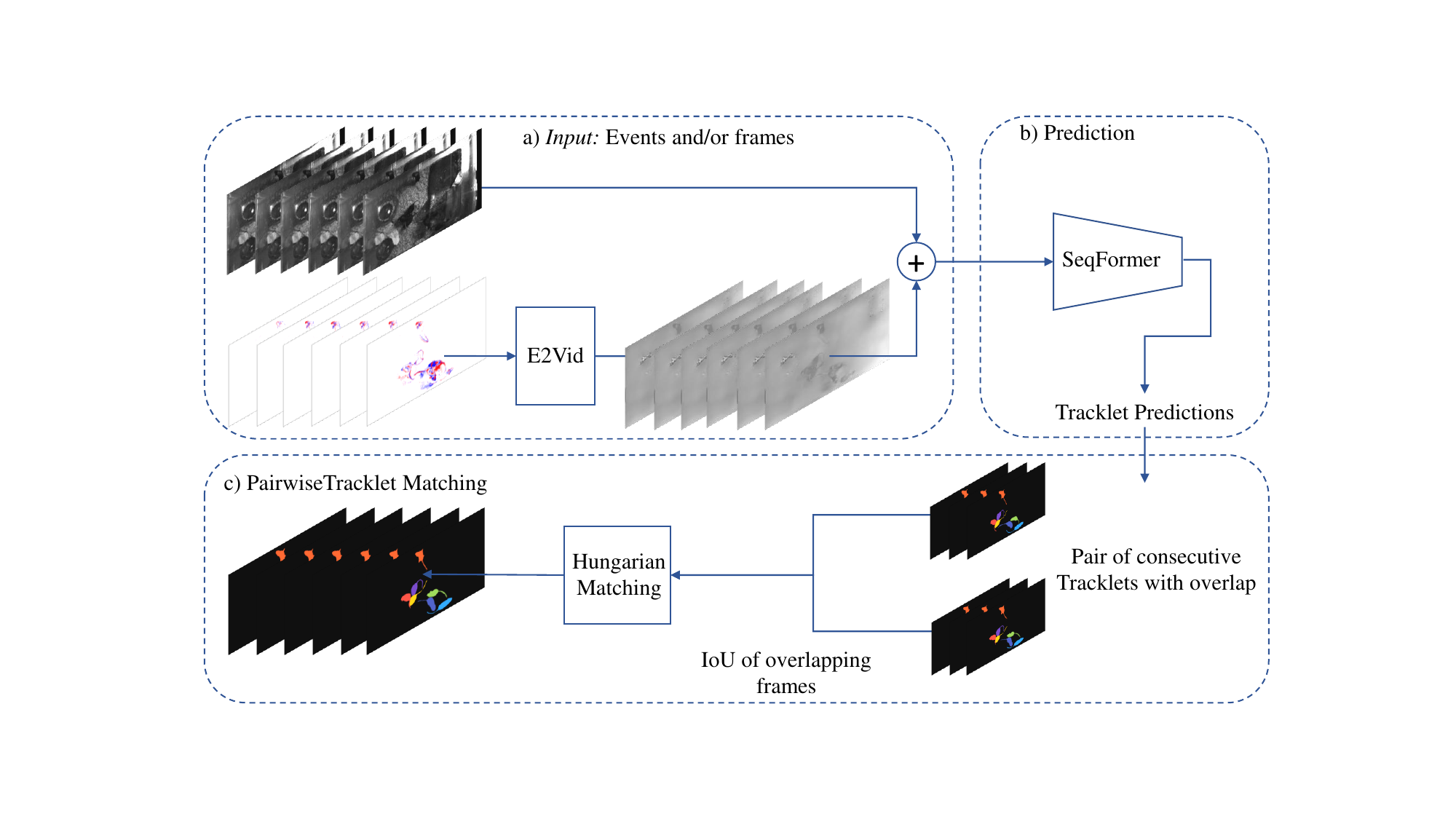}}
	\caption{\emph{Method 2 overview (EventSeqFormer).} 
 This method uses a tracking-by-query approach. 
 (a) We input an entire sequence of frames concatenated with E2VID images and divide them into smaller chunks with overlaps.
 (b) Within each chunk, we input the frames into SeqFormer for inference resulting in tracklets. 
 (c) Due to the 20-frame overlap between each tracklet, we associate each tracklet by performing Hungarian matching based on the IoU of instances, ultimately producing a tracking result for the entire sequence.
 \label{fig:method:seqformer}}
\end{figure*}

In addition to the tracking-by-detection method, we provide a tracking-by-query approach based on the SeqFormer architecture~\cite{Wu22eccv}, using transformers to model temporal and spatial dependencies for space-time instance segmentation.
A convolutional neural network (CNN) backbone and a transformer encoder extract per-frame feature maps (\cref{fig:method:seqformer}).
A fixed number of instance queries model information of the objects across time.
Cross attention is applied between instance queries and the feature maps for each frame independently.
Afterwards, an updated instance token is obtained as a weighted sum of the per-frame tokens.
The instance queries and per-frame box queries are decoded into class probabilities, masks, and bounding boxes.
For a full description, we refer the reader to \cite{Wu22eccv}.

To use the architecture with events, we use two strategies.
In one case we reconstruct grayscale frames at the timestamps using E2VID~\cite{Rebecq19pami}, in the second we use voxel grids, where the events within two consecutive frame timestamps are divided into several time bins and each resulting spacetime voxel carries the number of signed events falling within it.

We train several versions of the model, one with grayscale frames and event representation combined, and one each with the respective single modality.
During inference, the SeqFormer model produces multiple predictions, with each prediction corresponding to an instance tracked over the selected frames. To select the most reliable predictions, we retained the top 30 predictions based on the scores and then applied non-maximum suppression to filter out redundant masks.

We perform inference at full spatial resolution.
Due to GPU memory constraints, we could process a maximum of 140 images simultaneously.
Therefore, we adopted a sliding-window approach to handle the entire video, which consists of approximately 600 frames.
The window size was set to 140 frames with a step size of 120 frames, creating a 20-frame overlap between consecutive windows.
Within these overlapping regions, instance association was performed using Hungarian matching, where instances were linked based on the mean intersection over union (IoU).
If the mean IoU exceeded a predefined threshold, instances were considered the same; otherwise, they were treated as new instances. 

\section{Experiments}

\subsection{Metrics}

We report several metrics to quantify the performance of our tracker, namely, the mask versions of Multi-Object Tracking Accuracy (MOTA) from the CLEAR metrics~\cite{Bernardin08jivp}, the IDF1 metric~\cite{Ristani16eccv} and the Higher Order Tracking Accuracy (HOTA)~\cite{Luiten21ijcv}.

MOTA emphasizes detection performance, IDF1 focuses on identity preservation ability, and HOTA balances the effects of accurate association, detection, and localization.
We also report DetA and AssA factors of the HOTA metric, which allow for evaluating detection and association performance separately.

\subsection{Results}

\subsubsection{Quantitative.}
\Cref{tab:exp:tracking_results} presents the results of our two methods.
Overall, the SORT-based method using both modalities and the SeqFormer-based model using only frames perform best, with a slight margin of 1\% in HOTA between them.
Comparing these two, the DetA and AssA metrics reveal that the SORT-based approach has a 6\% advantage in detection accuracy, while SeqFormer shows a 5\% better association accuracy. 
This better association accuracy of the end-to-end learned method is expected, as SORT performs frame-by-frame matching, whereas SeqFormer considers a longer context window.

\Cref{tab:exp:results_per_sequence} displays results for each test set sequence.
Notably, sequences 1 and~7 have a low contrast sensitivity (event camera threshold), which E2VID does not generalize well to.
Consequently, the event-only models using E2VID frames as intermediate representations fail on these sequences, affecting overall performance.
Furthermore, it can be seen that for the \mname{} method, models utilizing event information consistently outperform the frame-only model, indicating that event data indeed aids the tracking task.

For \mnametwo, performance between the different modalities varies across sequences.
This suggests that better domain adaptation methods are necessary to fully leverage the transformer architecture's capabilities using event information. Lastly, \cref{tab:exp:sensitivity} presents a sensitivity study for the NMS thresholds of each method.

In summary, the results indicate that events can improve tracking accuracy. 
However, the dataset poses challenges, as pre-trained components like E2VID fail under one of the contrast threshold settings.
Improved integration of event and frame information could further amplify the advantages of event-based tracking.

\begin{table}[t]
\centering
\caption{Comparison of the tracker on different modalities.}
\label{tab:exp:tracking_results}
\adjustbox{max width=\linewidth}{
\setlength{\tabcolsep}{4pt}
\begin{tabular}{l*{8}{S[table-format=2.3]}}
\toprule
 \textbf{Method} &
   \textbf{\text{Frames}}       & 
   \textbf{\text{Events}}       & 
   {\textbf{MOTA} $\uparrow$}   & 
   {\textbf{IDF1} $\uparrow$}   & 
   {\textbf{HOTA} $\uparrow$}   & 
   {\textbf{DetA} $\uparrow$}   & 
   {\textbf{AssA} $\uparrow$}   \\
\midrule

 \mname                &
   \cmark              & 
   \xmark              & 
    34.42              & 
    45.41              & 
    41.83              &
    46.67              &
    38.451             \\

 \mname                &
   \xmark              & 
   \cmark              & 
    32.127             & 
    40.064             & 
    33.684             &
    33.575             &
    34.073             \\

 \mname                &
   \cmark              & 
   \cmark              & 
    \bnum{54.937}      & 
    \bnum{65.174}      & 
    \bnum{54.188}      &
    \bnum{53.694}      &
    \unum{2.2}{55.912} \\

 \mnametwo             &
   \cmark              & 
   \xmark              & 
    \unum{2.2}{40.216} & 
    \unum{2.2}{61.417} & 
    \unum{2.2}{53.073} &
    \unum{2.2}{47.568} &
    \bnum{60.271}      \\

 \mnametwo~(E2VID)     &
   \xmark              & 
   \cmark              & 
    -16.338            &
    34.82              &
    30.517             & 
    24.262             &
    38.582             \\

 \mnametwo~(E2VID)     &
   \cmark              & 
   \cmark              & 
    39.454             & 
    56.119             & 
    47.359             &
    45.656             &
    49.567             \\
    
 \mnametwo~(Voxel)     &
   \cmark              & 
   \cmark              & 
    40.72              & 
    60.139             & 
    47.824             &
    44.232             &
    52.406             \\

\mnametwo~(Voxel)      &
   \xmark              & 
   \cmark              & 
    -67.979            & 
    24.908             & 
    23.136             &
    16.491	           &
    32.627             \\

\bottomrule
\end{tabular}
}
\end{table}

\begin{table}[t]
\centering
\caption{HOTA per sequence for several trackers.
\emph{Contrast threshold} indicates which set of contrast thresholds was used to record the corresponding sequence: 
L for the low and H for the high setting, according to the bias values in \cref{tab:data:statistics}.
}
\label{tab:exp:results_per_sequence}
\adjustbox{max width=\linewidth}{
\setlength{\tabcolsep}{3pt}
\begin{tabular}{l*{12}{S[table-format=2.2]}}
\toprule

 \textbf{Method}  &
 \textbf{\text{Frames}}  &
 \textbf{\text{Events}}  &
   \textbf{1}   & 
   \textbf{7}   & 
   \textbf{10}  & 
   \textbf{16}  & 
   \textbf{22}  & 
   \textbf{26}  & 
   \textbf{28}  & 
   \textbf{32}  &
   \textbf{\text{Combined}}  &
   \textbf{\text{Combined}}\\
\cmidrule{1-11}
  Contrast threshold  &
             &
             &
   L         & 
   L         &  
   H         & 
   H         & 
   H         & 
   H         & 
   H         & 
   H         &
   \textbf{\text{w/o 1 \& 7}}  &
             \\
\cmidrule{1-11}

  \mname     &
   \cmark    &
   \xmark    &
   33.597    & 
   59.247    & 
   41.510    & 
   39.902    & 
   30.686    & 
   27.457    & 
   29.565    & 
   55.021    &
   38.489    &
   41.833   \\

 \mname     &
   \xmark   &
   \cmark   &
    0       & 
    0       & 
   21.405   & 
   \bnum{49.23}    & 
   \bnum{40.367}   & 
   32.186   & 
   \bnum{55.169}   & 
   52.712   &
   \unum{2.2}{40.609}   &
   33.684  \\

 \mname   &
   \cmark       &
   \cmark       &
   59.737       & 
   \unum{2.2}{78.126}    & 
   \unum{2.2}{45.299}    & 
   \unum{2.2}{46.077}    & 
   33.441                & 
   \unum{2.2}{36.317}    & 
   \unum{2.2}{45.876}    & 
   61.322                &
   \bnum{45.598} &
   \bnum{54.188}        \\

  \mnametwo  &
   \cmark    &
   \xmark    &
    \bnum{84.758}   & 
    \bnum{79.299}   & 
    \bnum{50.876}   & 
    22.479          & 
    35.089          & 
    27.433          & 
    22.569          & 
    \unum{2.2}{65.744} &
    39.734 &
    \unum{2.2}{53.073} \\

 \mnametwo~(E2VID)  &
   \xmark      &
   \cmark      &
    2.279      & 
    2.2084     & 
    39.367     & 
    21.23      & 
    30.402     & 
    \bnum{42.162} & 
    37.07      & 
    55.619     &
    38.95 &
    30.517    \\

 \mnametwo~(E2VID)   &
   \cmark       &
   \cmark       &
   \unum{2.2}{68.255}  & 
   60.53        & 
   42.12        & 
   27.67        & 
   \unum{2.2}{39.21}  & 
   30.301       & 
   28.816       & 
   \bnum{66.16} &
   39.982       &
   47.359      \\

 \mnametwo~(Voxel)      &
   \cmark       &
   \cmark       &
   63.393  & 
   65.361        & 
   47.515        & 
   24.563        & 
   31.831  & 
   26.018       & 
   32.028       & 
   65.312 &
40.332      &
   47.824      \\

 \mnametwo~(Voxel) &
   \xmark          &
   \cmark          &
   2.5467          & 
   7.6314          & 
   28.771          & 
   21.049          & 
   21.793          & 
   37.905          & 
   23.066          & 
   53.474          &
   31.919          &
   23.136          \\

\bottomrule
\end{tabular}
}
\end{table}

\begin{table}[t]
\centering
\caption{Sensitivity of non-maximum suppression (NMS) thresholds.
Please note that both methods use non-maximum suppression, but in different contexts.}
\label{tab:exp:sensitivity}
\adjustbox{max width=\linewidth}{
\setlength{\tabcolsep}{6pt}
\begin{tabular}{ll*{3}{S[table-format=2.3]}}
\toprule

\multirow{2}{*}{\textbf{Method}} & \multicolumn{3}{c}{\(\sigma\)} \\
\cmidrule(lr){2-4}
& {0.1} & {0.3} & {0.5} \\
\midrule

 \mname & 54.188 & 53.855 & 51.731 \\

\midrule

 \mnametwo & 47.359 & 42.562 & 37.225 \\
\midrule
 \mnamethree & 47.82 &43.054 & 37.463 \\

\bottomrule
\end{tabular}
}
\end{table}

\def\figWidth{0.236\linewidth}
\begin{figure*}[t!]
	\centering
    {\scriptsize
    \setlength{\tabcolsep}{1pt}
	\begin{tabular}{
	>{\centering\arraybackslash}m{0.3cm} 
	>{\centering\arraybackslash}m{\figWidth} 
	>{\centering\arraybackslash}m{\figWidth} 
	>{\centering\arraybackslash}m{\figWidth} 
	>{\centering\arraybackslash}m{\figWidth}
    }
		\rotatebox{90}{\makecell{Frames}}
		&\includegraphics[clip,trim={0cm 0cm 0cm 0cm},width=\linewidth]{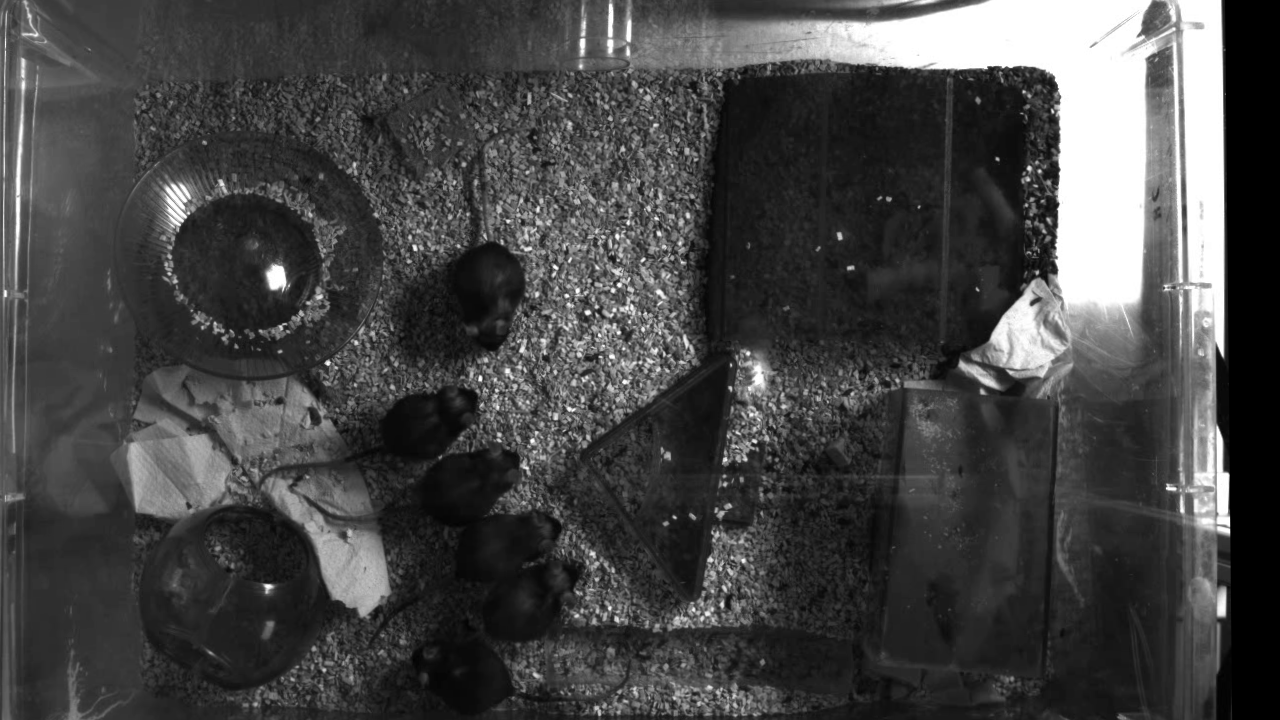}
		&\includegraphics[clip,trim={0cm 0cm 0cm 0cm},width=\linewidth]{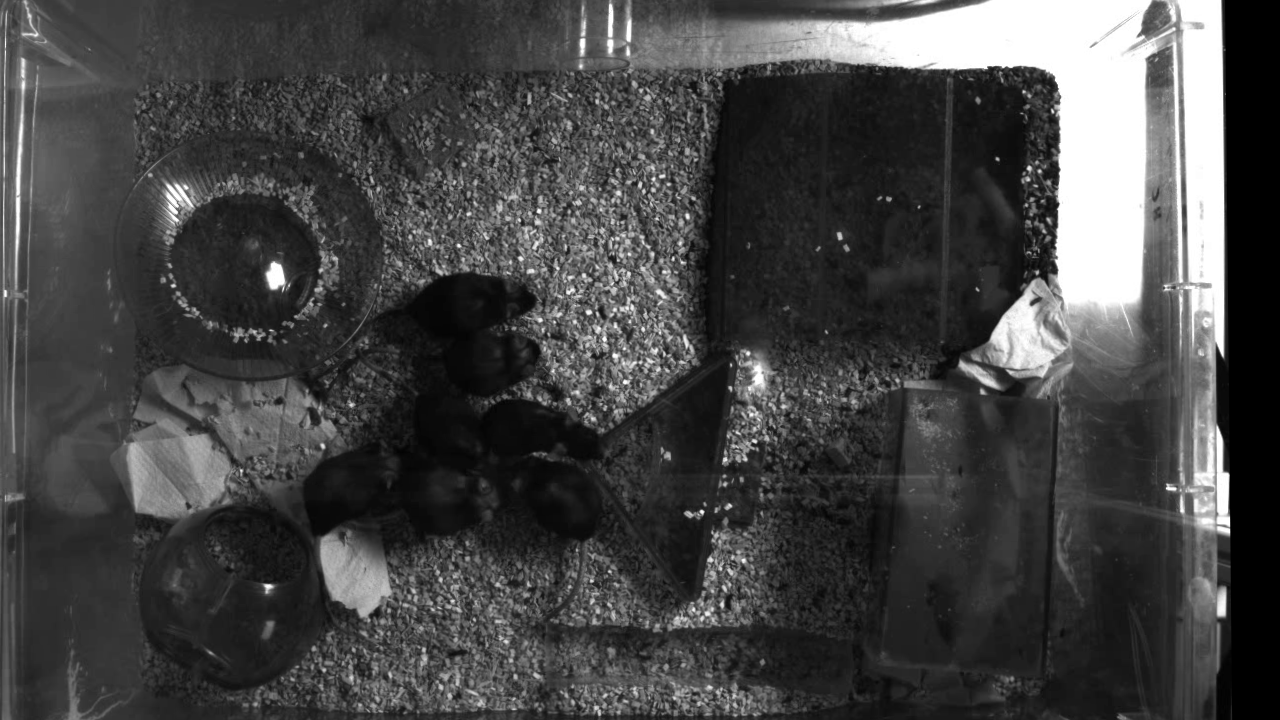}
		&\includegraphics[clip,trim={0cm 0cm 0cm 0cm},width=\linewidth]{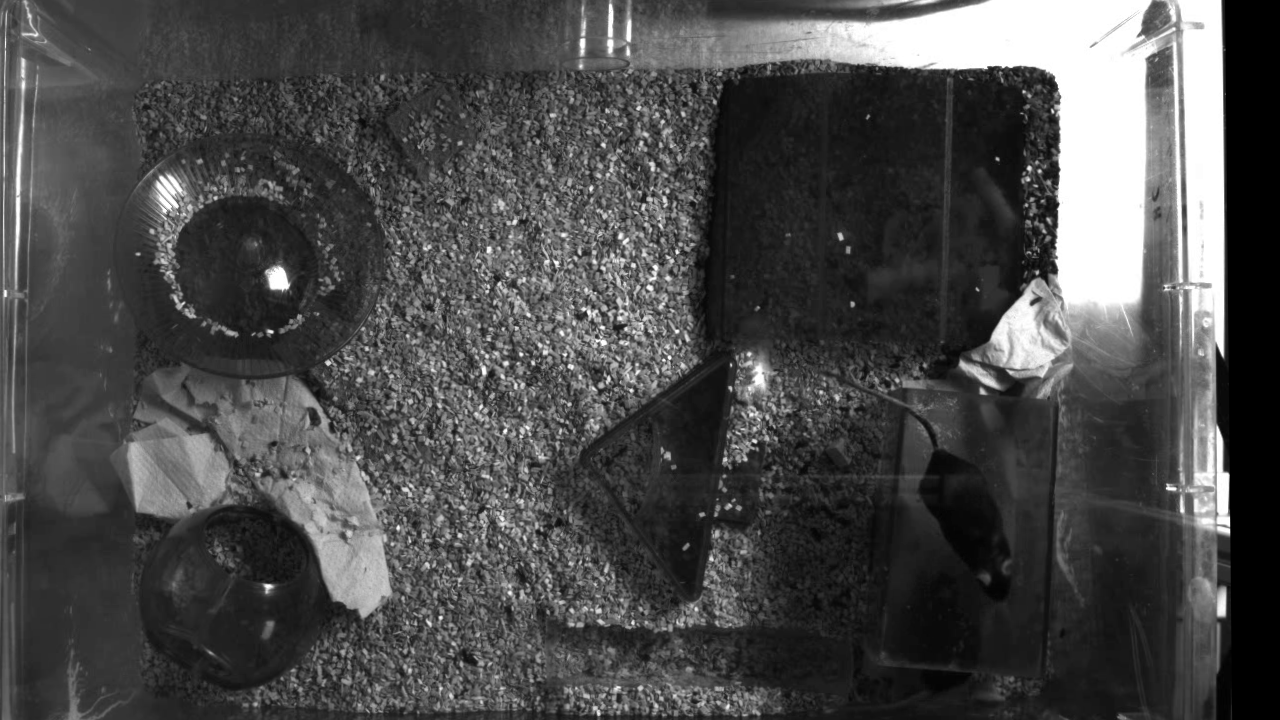}
		&\includegraphics[clip,trim={0cm 0cm 0cm 0cm},width=\linewidth]{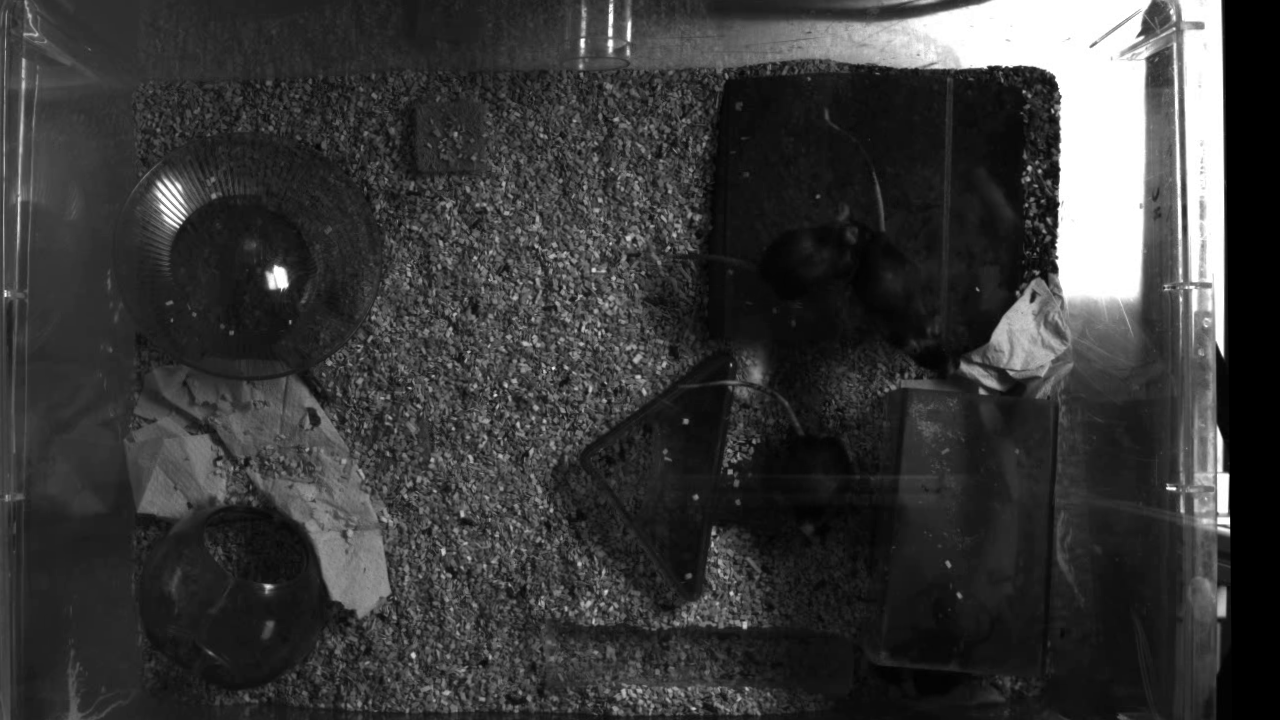} \\
		\rotatebox{90}{\makecell{Events}}
		&\gframe{\includegraphics[clip,trim={0cm 0cm 0cm 0cm},width=\linewidth]{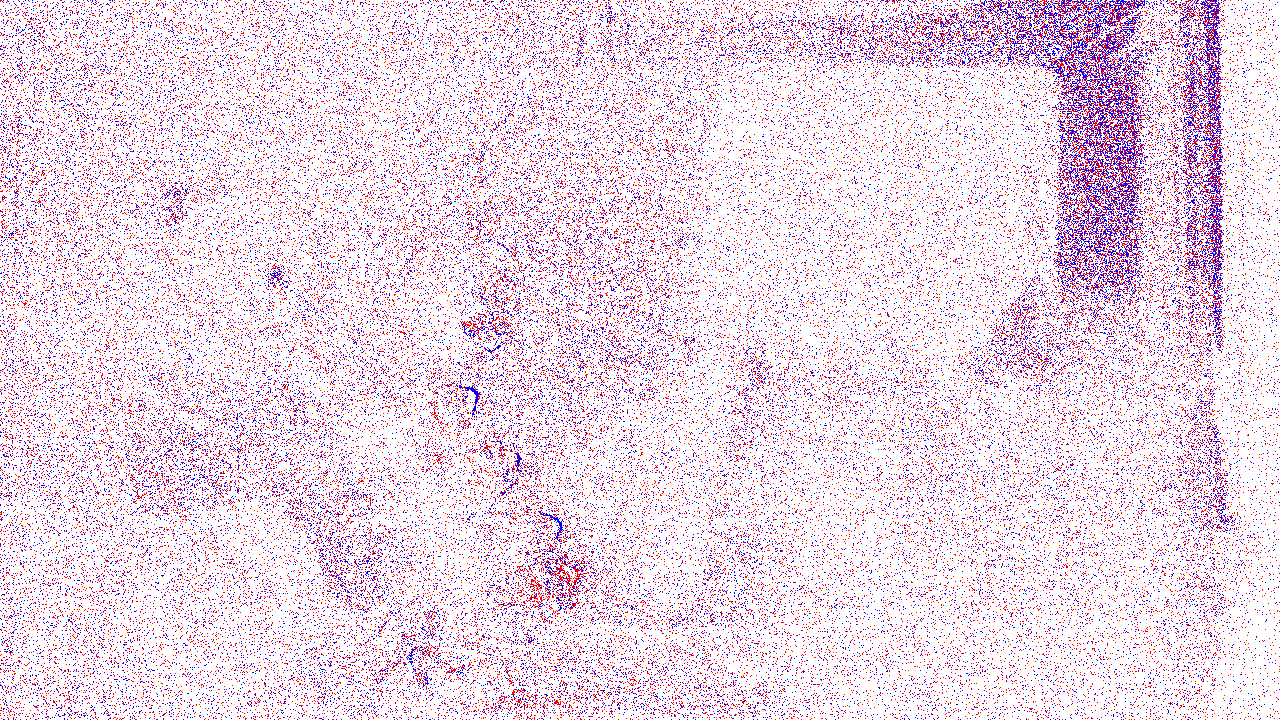}}
		&\gframe{\includegraphics[clip,trim={0cm 0cm 0cm 0cm},width=\linewidth]{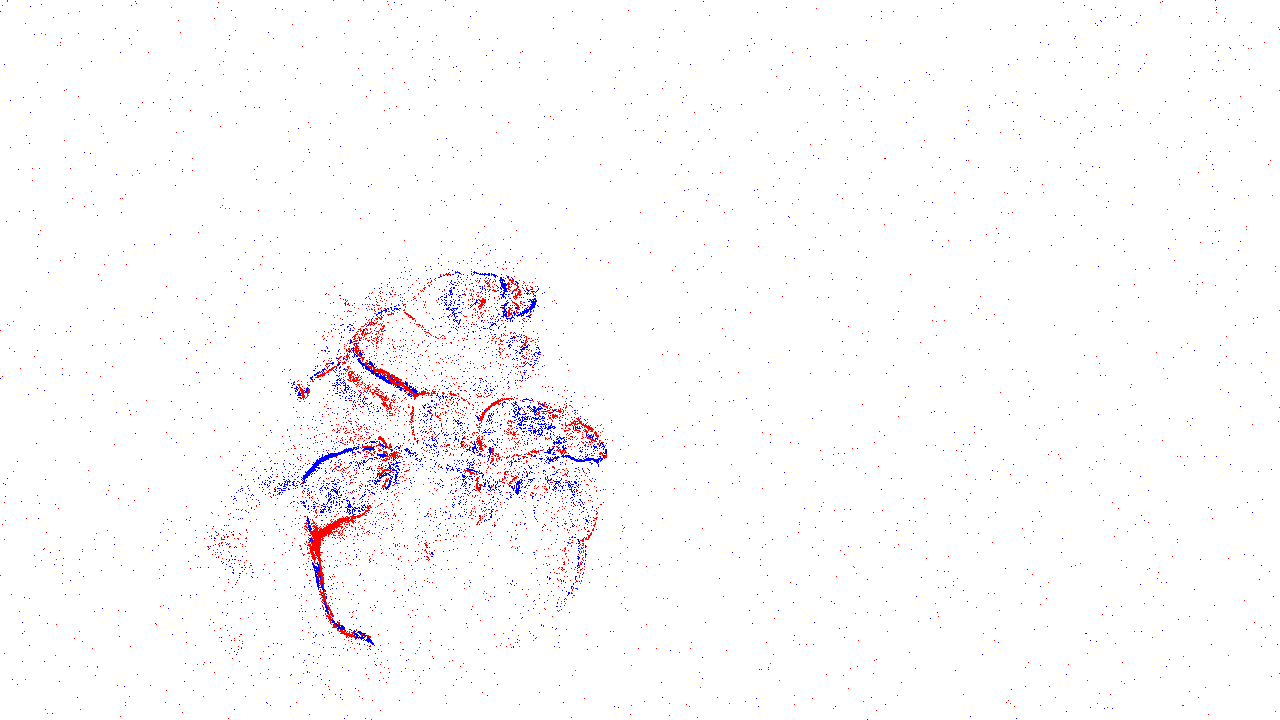}}
		&\gframe{\includegraphics[clip,trim={0cm 0cm 0cm 0cm},width=\linewidth]{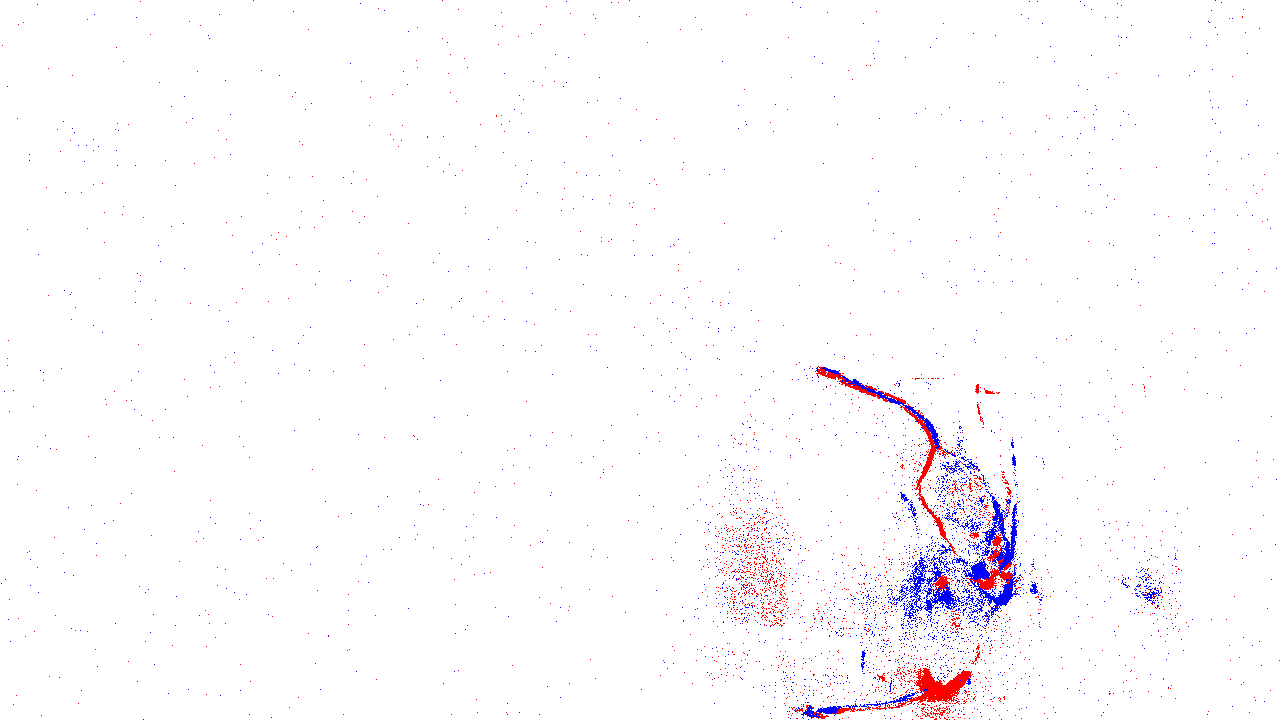}}
		&\gframe{\includegraphics[clip,trim={0cm 0cm 0cm 0cm},width=\linewidth]{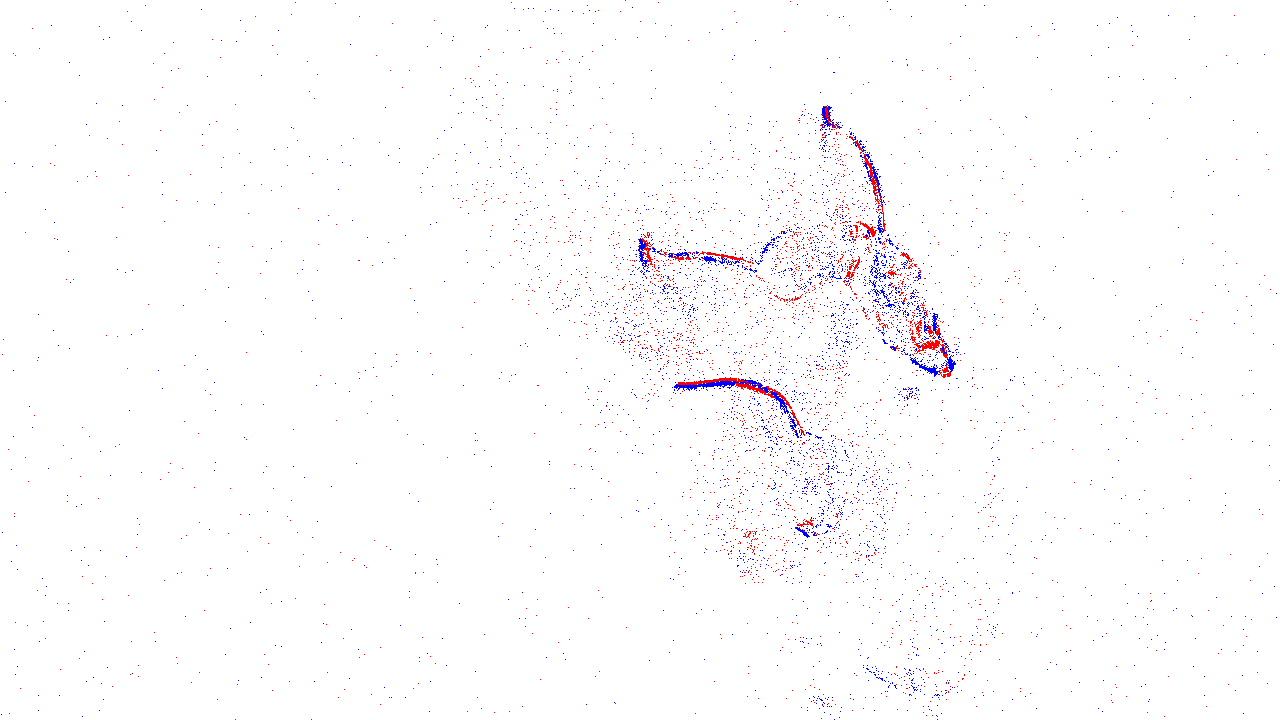}} \\
		\rotatebox{90}{\makecell{GT}}
		&\includegraphics[clip,trim={0cm 0cm 0cm 0cm},width=\linewidth]{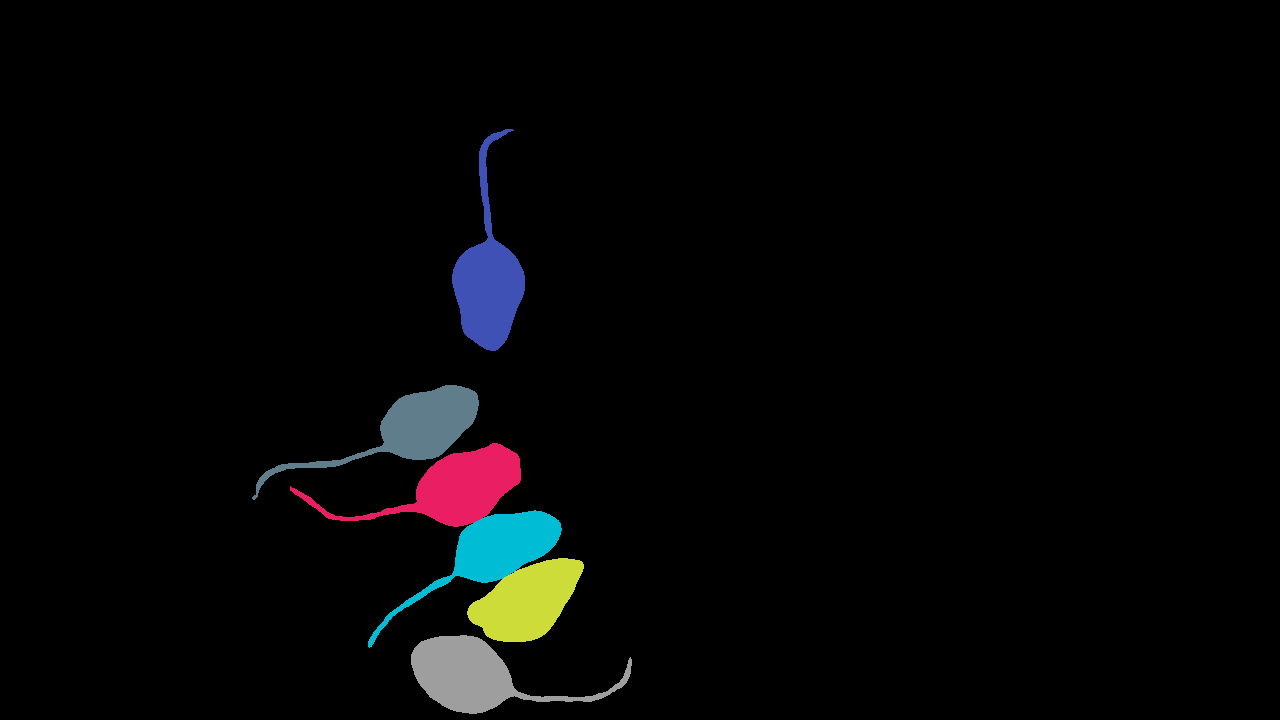}
		&\includegraphics[clip,trim={0cm 0cm 0cm 0cm},width=\linewidth]{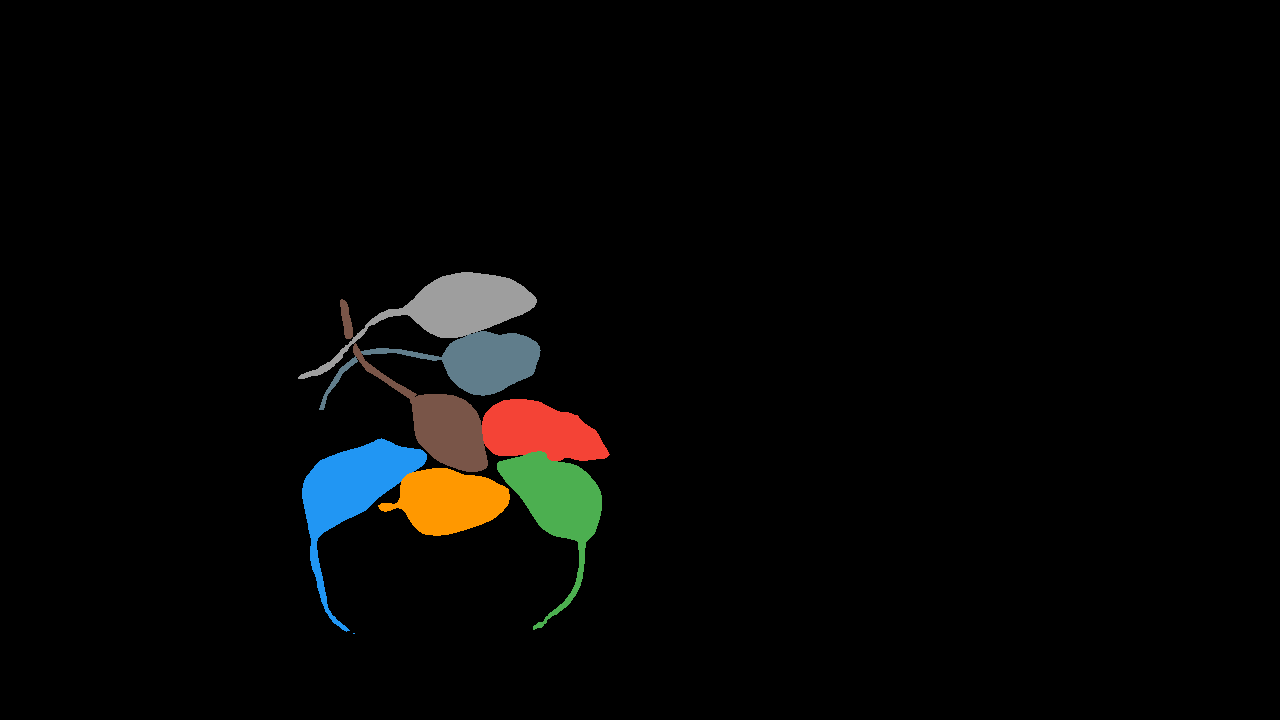}
		&\includegraphics[clip,trim={0cm 0cm 0cm 0cm},width=\linewidth]{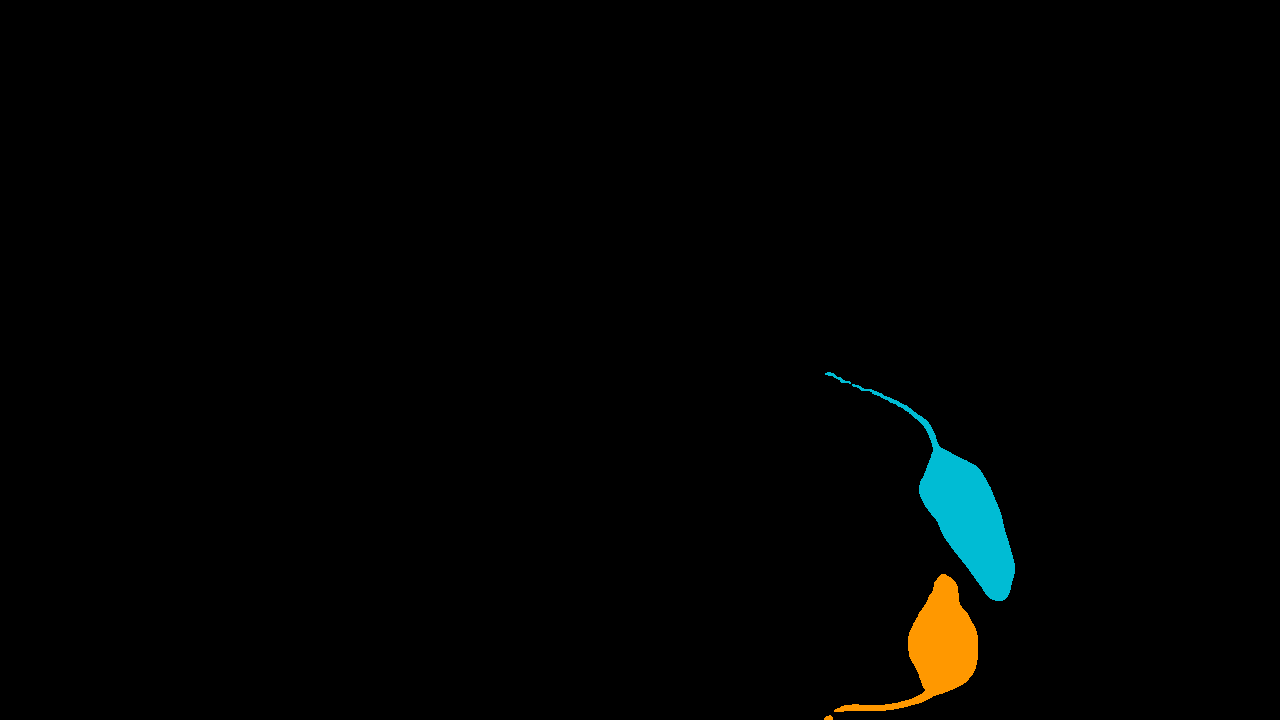}
		&\includegraphics[clip,trim={0cm 0cm 0cm 0cm},width=\linewidth]{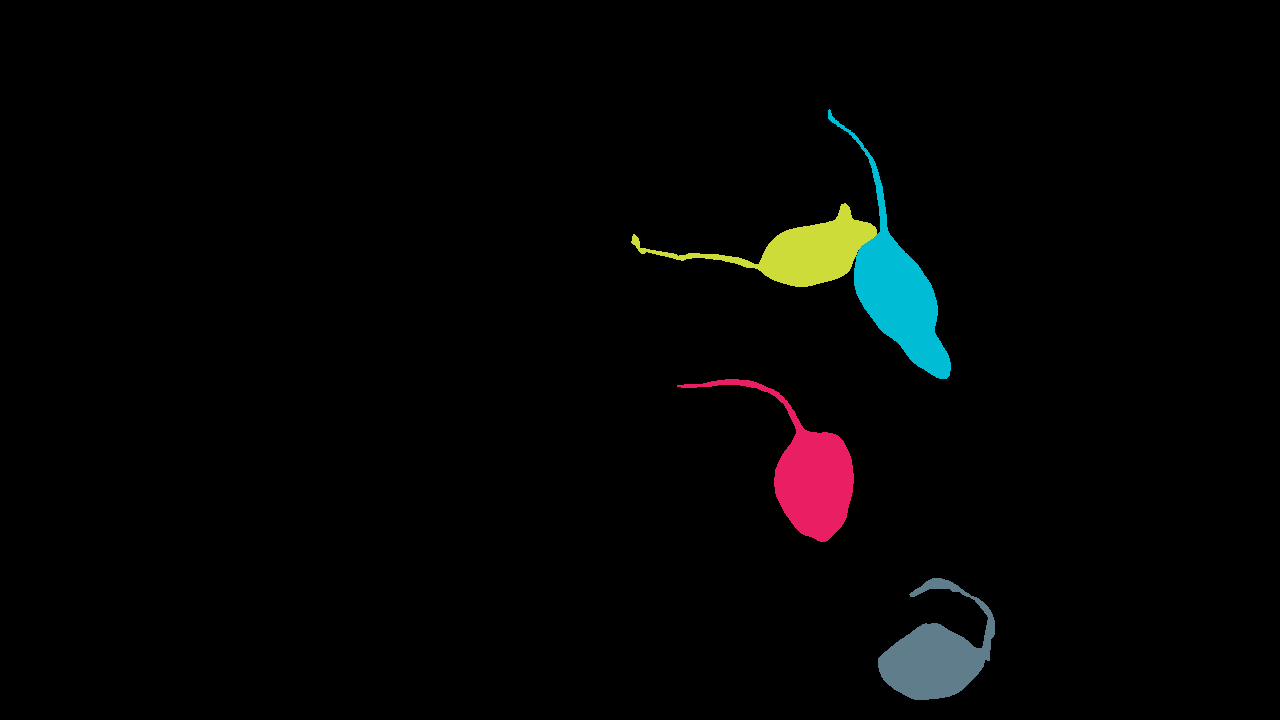} \\
		\rotatebox{90}{\makecell{MMSort}}
		&\includegraphics[clip,trim={0cm 0cm 0cm 0cm},width=\linewidth]{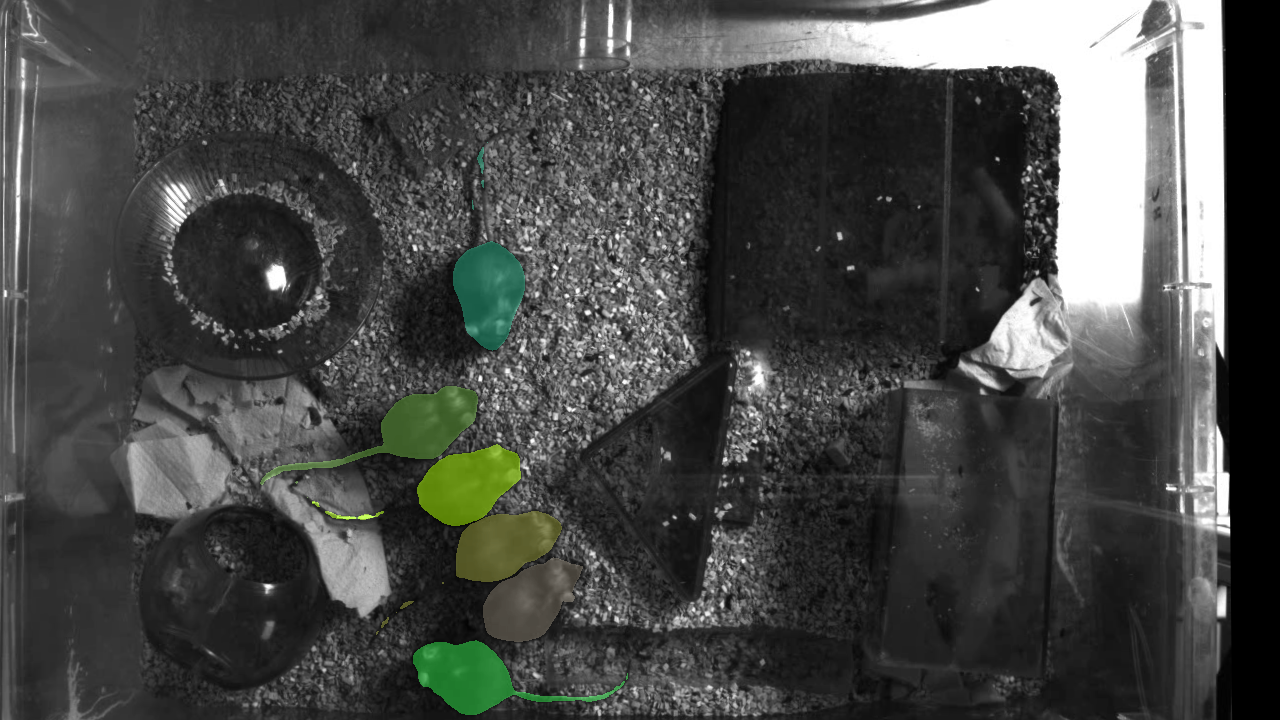}
		&\includegraphics[clip,trim={0cm 0cm 0cm 0cm},width=\linewidth]{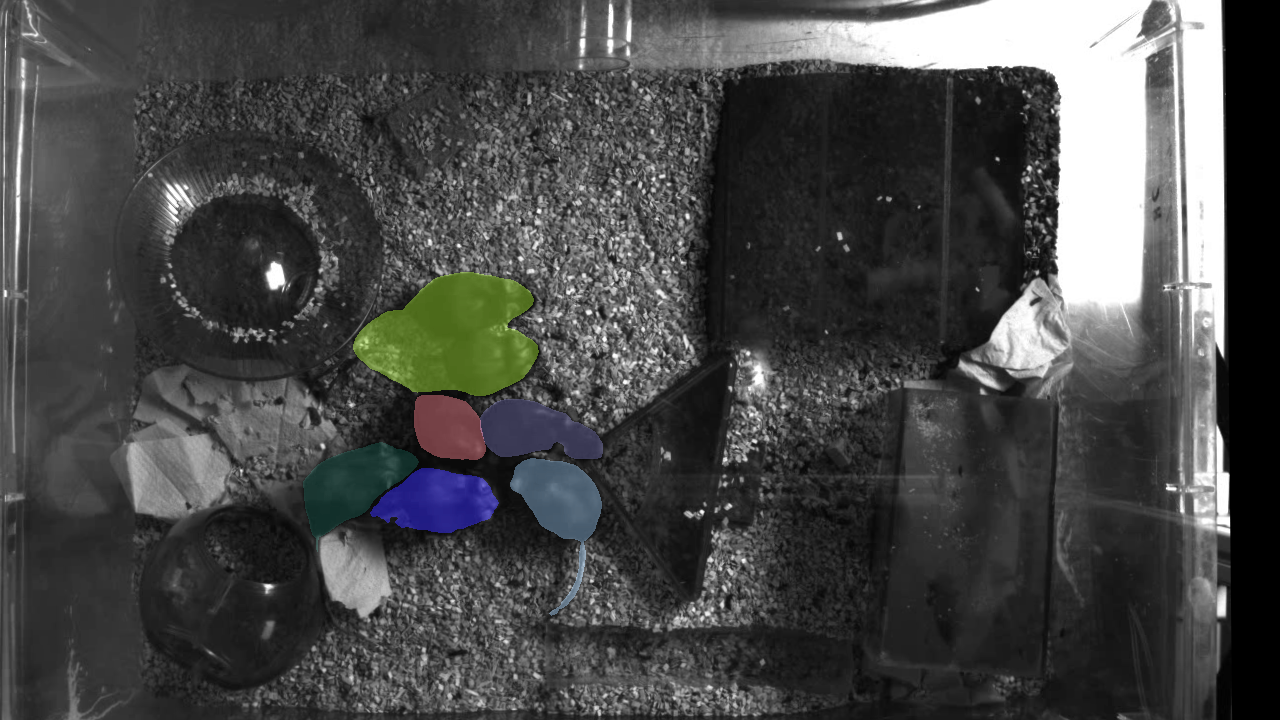}
		&\includegraphics[clip,trim={0cm 0cm 0cm 0cm},width=\linewidth]{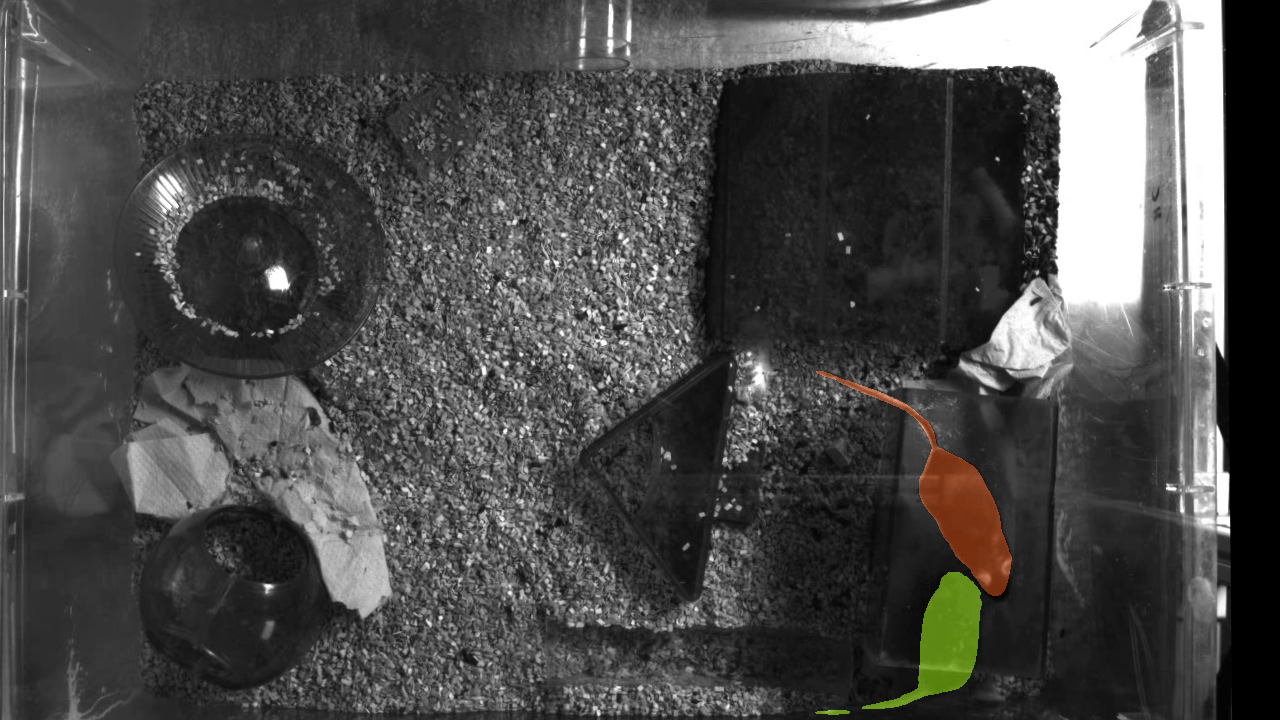}
		&\includegraphics[clip,trim={0cm 0cm 0cm 0cm},width=\linewidth]{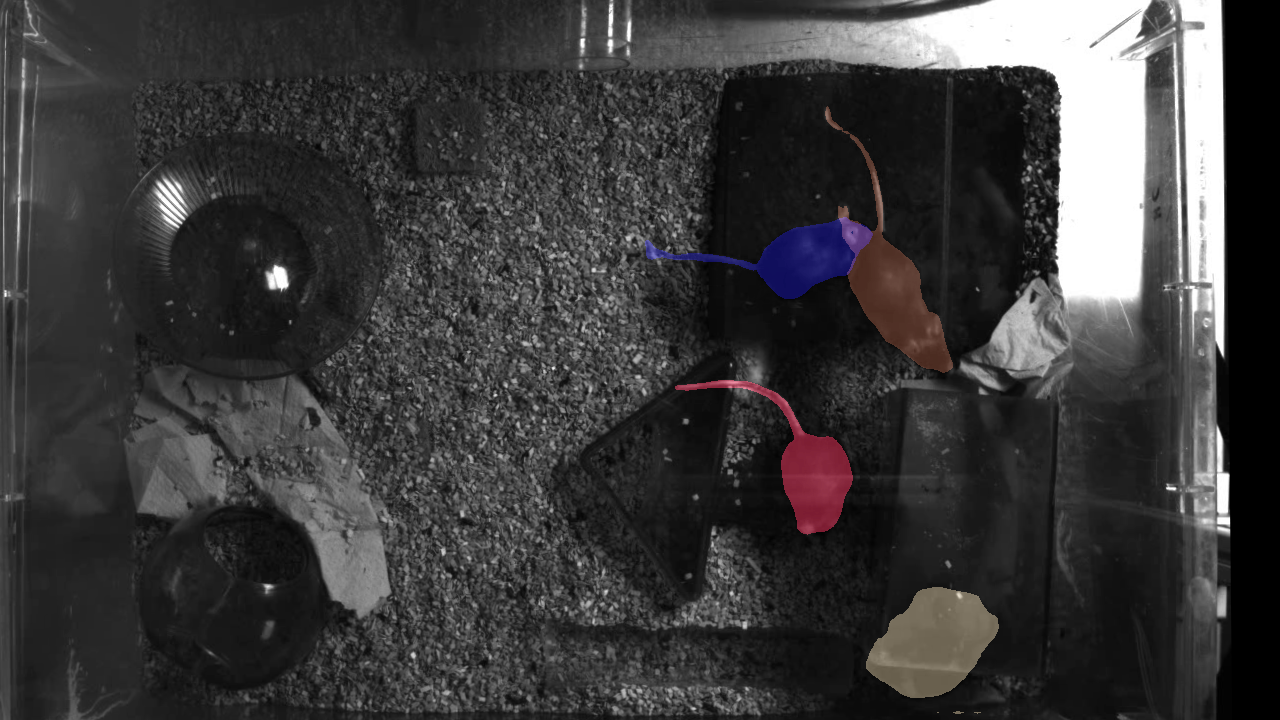} \\
		\rotatebox{90}{\makecell{ESeqFormer}}
		&\includegraphics[clip,trim={0cm 0cm 0cm 0cm},width=\linewidth]{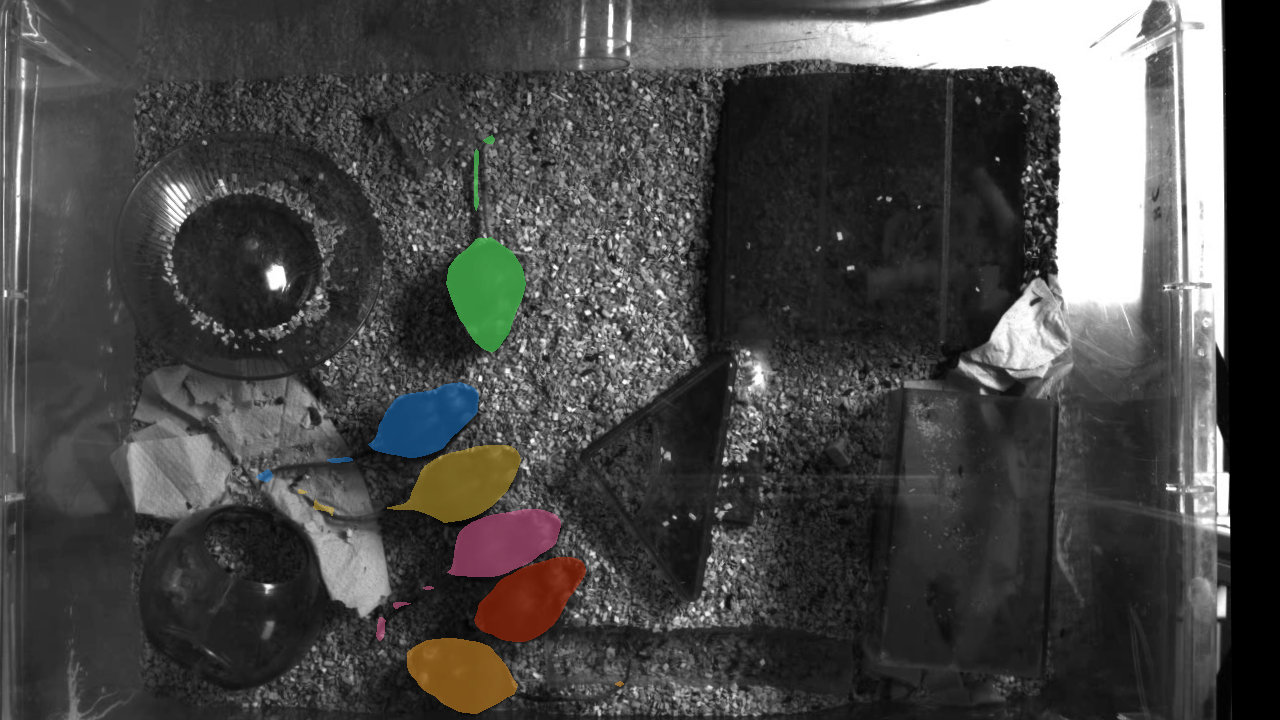}
		&\includegraphics[clip,trim={0cm 0cm 0cm 0cm},width=\linewidth]{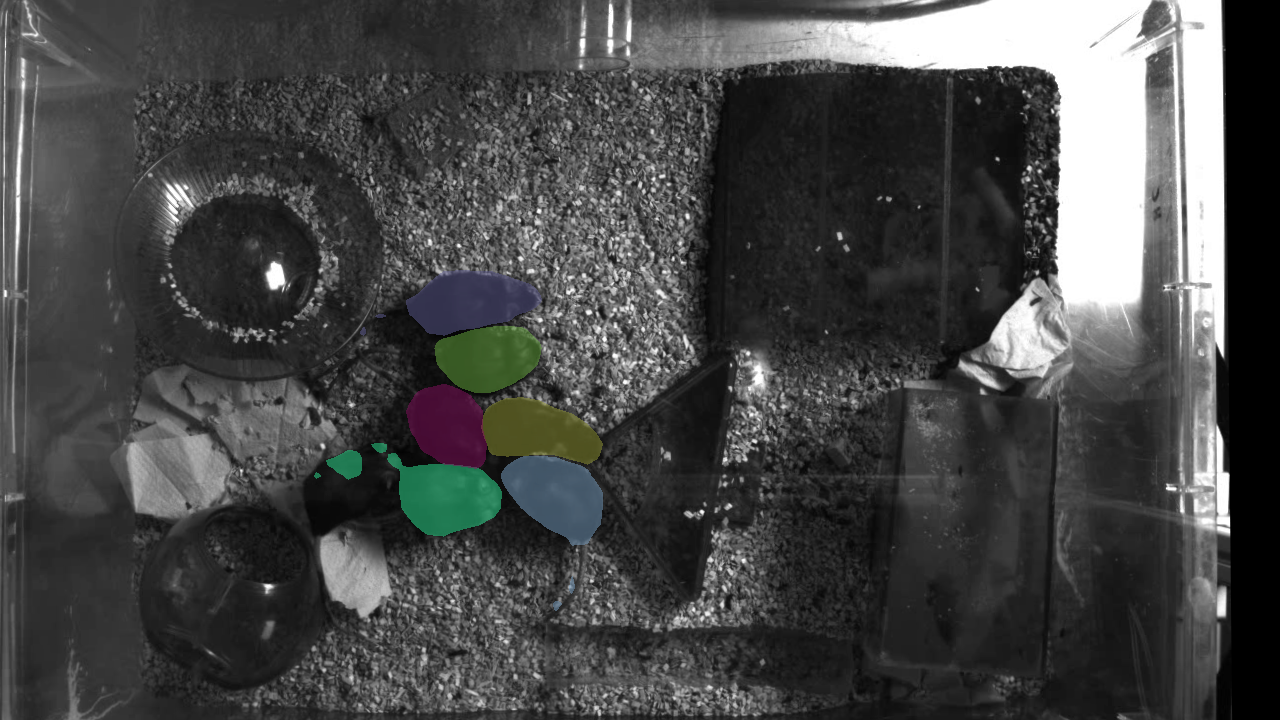}
		&\includegraphics[clip,trim={0cm 0cm 0cm 0cm},width=\linewidth]{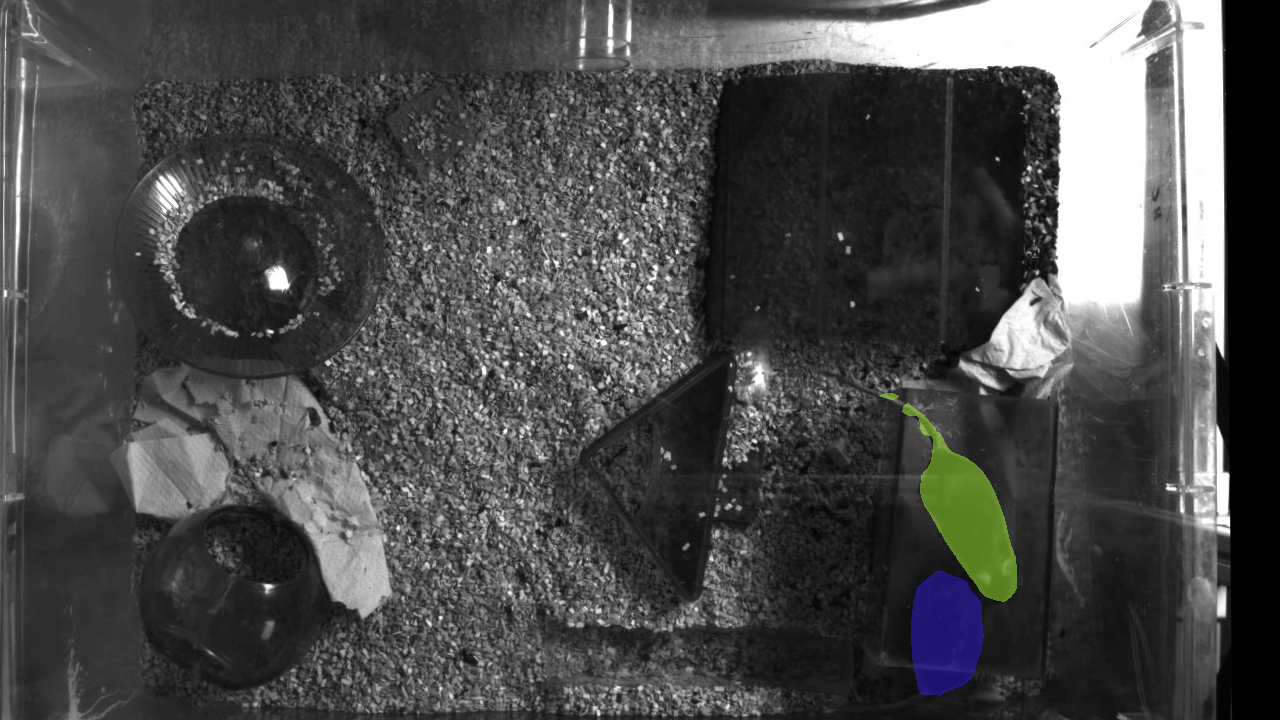}
		&\includegraphics[clip,trim={0cm 0cm 0cm 0cm},width=\linewidth]{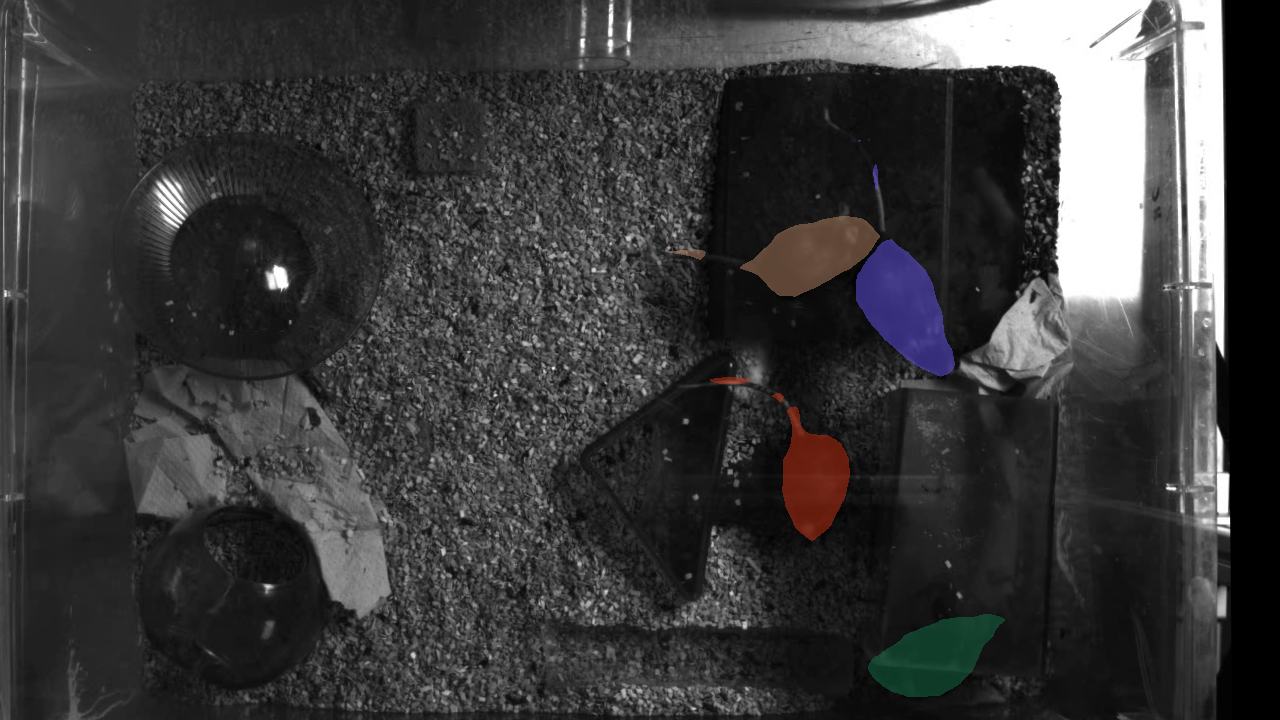} \\
        & a) Low thresholds
        & b) Seven instances
        & c) Partial occlusion
        & d) Partial occlusion\\
	\end{tabular}
	}
	\caption{Sample ``snapshots'' of our dataset and the corresponding predictions by the test methods.}
	\label{fig:dataset_examples}
\end{figure*}

\subsubsection{Qualitative.}
\Cref{fig:dataset_examples} shows examples of the \dname{} dataset alongside the predictions of our two methods at the frame's timestamp.
Column a) shows the low-threshold setting resulting in high event-noise levels.
Column b) provides an example of the maximum number of target instances (seven), with them crowded together.
A common failure case of \mname~in this setting is a wrong segmentation by SAM.
Columns c) and d) present cases of partial occlusion.
The houses in the cage are somewhat transparent but instances below the roof are hard to detect.

\section{Limitations}

While our dataset provides challenging conditions, all sequences show the same general scene and one instance class (albeit deformable).
The aligned frames and events make \dname{} perfectly suited to benchmark frames- against event-based methods and to develop methods truly using the nature of event data for tracking, the usage for the development of generalizing methods is limited by the lack of scene variety.

The two reference methods show the potential of the event data.
However, both rely on E2VID, a pre-trained network, and therefore fail when E2VID fails, like in our high-noise case.
SAM has no notion of a class, therefore \mname~has a common failure case, where the box prompt has the correct positions but the derived binary mask is capturing wrong objects like the two mice in \cref{fig:dataset_examples} b). 
While adapting off-the-shelf methods can provide baseline results, dedicated space-time instance segmentation methods are yet to be explored, which is nevertheless enabled and encouraged by our work.

\section{Conclusion}

We introduced the first space-time instance segmentation (SIS) dataset.
It consists of 33 clips of about 20s with aligned frames and events.
The content is a mouse cage with up to seven mice instances featuring occlusions, deformations, and uneven illumination conditions.
Alongside our \dname{} dataset, we provide two reference methods and an extensive evaluation of the dataset.
The first method is based on the tracking-by-detection paradigm loosely coupling several pre-trained models and a fine-tuned object detector.
The second is end-to-end learned, and based on a recent frame-based video instance segmentation method.
The results show the potential of using events for tracking, while also highlighting the challenges of the new dataset and opportunities for further improvements on our dataset.
We hope \dname{} opens new avenues for event-based scene understanding.

\FloatBarrier

\section*{Acknowledgments}
Funded by the Deutsche Forschungsgemeinschaft (DFG, German Research Foundation) under Germany’s Excellence Strategy -- EXC 2002/1 ``Science of Intelligence'' -- project number 390523135.
We thank Mr. Berin Balachandar and the team at AnnotateX Pte Ltd for their valuable support in the annotation process.

\end{document}